\documentclass{article}
\newcommand{\keywords}[1]{\textbf{Keywords:}\quad #1}

\usepackage{longtable}
\usepackage{multirow} 
\usepackage{booktabs} 
\usepackage{rotating} 
\usepackage[left=1in, right=1in, top=1in, bottom=1in]{geometry}
\usepackage{graphicx}
\graphicspath{{images/}}

\usepackage{lmodern} 
\usepackage{graphicx} 
\usepackage{subfig} 
\usepackage{lscape}
\usepackage{hyperref}
\hypersetup{
	colorlinks=true,
	citecolor=blue,
	linkcolor=red,
	filecolor=magenta,      
	urlcolor=blue
}
\usepackage{verbatim} 
\usepackage{authblk}

\title{Handwritten Optical Character Recognition (OCR): A Comprehensive Systematic Literature Review (SLR) }

\author[1,4]{Jamshed Memon \thanks{All authors have contributed equally}}
\author[3]{Maira Sami}
\author[1,2]{Rizwan Ahmed Khan}

\affil[1]{Faculty of IT, Barrett Hodgson University, Karachi, Pakistan.}
\affil[2]{LIRIS, Universit\'e Claude Bernard Lyon1, France.}
\affil[3]{CIS, NED University of Engineering and technology, Karachi, Pakistan.}
\affil[4]{School of Computing, Quest International University, Perak, Malaysia.}
\date{}

\begin{document}
\maketitle




\begin{abstract}

Given the ubiquity of handwritten documents in human transactions, Optical Character Recognition (OCR) of documents have invaluable practical worth. Optical character recognition is a science that enables to translate various types of documents or images into analyzable, editable and searchable data. During last decade, researchers have used artificial intelligence / machine learning tools to automatically analyze handwritten and printed documents in order to convert them into electronic format. The objective of this review paper is to summarize research that has been conducted on character recognition of handwritten documents and to provide research directions. In this Systematic Literature Review (SLR) we collected, synthesized and analyzed research articles on the topic of handwritten OCR (and closely related topics) which were published between year 2000 to 2018. We followed widely used electronic databases by following pre-defined review protocol. Articles were searched using keywords, forward reference searching and backward reference searching in order to search all the articles related to the topic. After carefully following study selection process 142 articles were selected for this SLR. This review article serves the purpose of presenting state of the art results and techniques on OCR and also provide research directions by highlighting research gaps.



\end{abstract}

\section{Introduction}
Optical character recognition (OCR) is a system that converts the input text into machine encoded format \cite{Tappert}. Today, OCR is helping not only in digitizing the handwritten medieval manuscripts \cite{Kumar2018}, but also helping in converting the typewritten documents into digital form \cite{Radwan2018}. This has made the retrieval of the required information easier as one doesn’t have to go through the piles of documents and files to search the required information. Organizations are satisfying the needs of digital preservation of historic data \cite{HistMed}, law documents \cite{Ashley2010}, educational persistence \cite{Zanibbi2012} etc. 

An OCR system depends mainly, on the extraction of features and discrimination / classification of these features (based on patterns). Handwritten OCR have received increasing attention as a subfield of OCR.  It is further categorized into offline system \cite{pathan2012recognition,parvez2013offline} and online system \cite{connell2001template} based on input data. The offline system is a static system in which input data is in the form of scanned images while in online systems nature of input is more dynamic and is based on the movement of pen tip having certain velocity, projection angle, position and locus point. Therefore, online system is considered more complex and advance, as it resolves overlapping problem of input data that is present in the offline system.

One of the earliest OCR system was developed in 1940s, with the advancement in the technology over the time, the system became more robust to deal with both printed and handwritten characters and this led to the commercial availability of the OCR machines. In 1965, advance reading machine ``IBM 1287'' was introduced at the ``world fair'' in New York \cite{Mori1992}. This was the first ever optical reader, which was capable of reading handwritten numbers.  During 1970s,  researchers focused on the improvement of response time and performance of the OCR system.  

The next two decades from 1980 till 2000, software system of OCR was developed and deployed in educational institutes, census OCR \cite{wilkinson1992first} and for recognition of stamped characters on metallic bar \cite{Kovacs-V1995}. In early 2000s, binarization techniques were introduced to preserve historical documents in digital form and provide researchers the access to these documents \cite{wolf2002text,gatos2004adaptive,he2005comparison,sari2002off}. Some of the challenges of binarization of historical documents was the use of nonstandard fonts, printing noise and spacing. In mid of 2000 multiple applications were introduced that were helpful for differently abled people. These applications helped these people in developing reading and writing skills.

In the current decade, researchers have worked on different machine learning approaches which include Support Vector Machine (SVM), Random Forests (RF), $k$ Nearest Neighbor ($k$NN), Decision Tree (DT) \cite{Mitchell, lorigo2006offline, Khan2019a} etc. Researchers combined these machine learning techniques with image processing techniques to increase accuracy of optical character recognition system. Recently researchers has focused on developing techniques for the digitization of handwritten documents, primarily based on deep learning \cite{DL} approach. This paradigm shift has been sparked due to adaption of cluster computing and GPUs and better performance by deep learning architectures \cite{66287}, which includes Recurrent Neural Networks (RNN), Convolutional Neural Network (CNN), Long Short-Term Memory (LSTM) networks etc.

This Systematic Literature Review (SLR) will not only serve the purpose of presenting literature in the domain of OCR for different languages but will also highlight research directions for new researcher by highlighting weak areas of current OCR systems that needs further investigation. 

This article is organized as follows. Section \ref{RM} discusses review methodology employed in this article. Review methodology section includes review protocol, inclusion and exclusion criteria, search strategy, selection process, quality assessment criteria and meta data synthesis of selected studies. Statistical data from selected studies is presented in Section \ref{SR}. Section \ref{RQ} presents research question and their motivation. Section \ref{classi} will discuss different classifications methods which are used for handwritten OCR. The section will also elaborate on structural and statistical models for optical character recognition. Section \ref{dataset} will present different databases (for specific language) which are available for research purpose. Section \ref{lang} will present research overview of language specific research in OCR, while Section \ref{future} will highlight research trends. Section \ref{conclusion} will summarize this review findings and will also highlight gaps in research that needs attention of research community.


\section{Review methods} \label{RM}

As mentioned above, this Systematic Literature Review (SLR) aims to identify and present literature on OCR by formulating research questions and selecting relevant research studies. Thus, in summary this review was:

\begin {enumerate}

\item	To summarize existing research work (machine learning techniques and databases) on different languages of handwritten character recognition systems.
\item		To highlight research weakness in order to eliminate them through additional research.  
\item		To identify new research areas within the domain of OCR. 
\end {enumerate}

We will follow strategies proposed by Kitchenham et al. \cite{kitchenham2010systematic}. Following proposed strategy, in subsequent sub-sections review protocol, inclusion and exclusion criteria, search strategy process, selection process and data extraction and synthesis processes are discussed.

\subsection{Review protocol}

Following the philosophy, principles and measures of the Systematic Literature Review (SLR) \cite{kitchenham2010systematic}, this systematic study was initialized with the development of comprehensive review protocol. This protocol identifies review background, search strategy, data extraction, research questions and quality assessment criteria for the selection of study and data analysis. 

The review protocol is what that creates a distinction between an SLR and traditional literature review or narrative review  \cite{kitchenham2010systematic}. It also enhances the consistency of the review and reduces the researchers' biasness. This is due to the fact that researchers have to present search strategy and the criteria for the inclusion of exclusion of any study in the review.

\subsection{Inclusion and exclusion criteria}\label{inc-exc}

Setting up an inclusion and exclusion criteria makes sure that only articles that are relevant to study are included. Our criteria includes research studies from journals, conferences, symposiums and workshops on the optical character recognition of English, Urdu, Arabic, Persian, Indian and Chinese languages. In this SLR, we considered studies that were published from January 2000 to December 2018. 

Our initial search based on the keywords only, resulted in 954 research articles related to handwritten OCRs of different languages (refer Figure \ref{fig:figure1} for compete overview of selection process). After thorough review of the articles we excluded the articles that were not clearly related to a handwritten OCR, but appeared in the search, because of keyword match. Additionally, articles were also excluded based on duplicity, non-availability of full text and whether the studies were related to any of our research questions.

\subsection{Search strategy}

\begin{figure*} [!htb]
	\centering
		\includegraphics [scale=0.85]{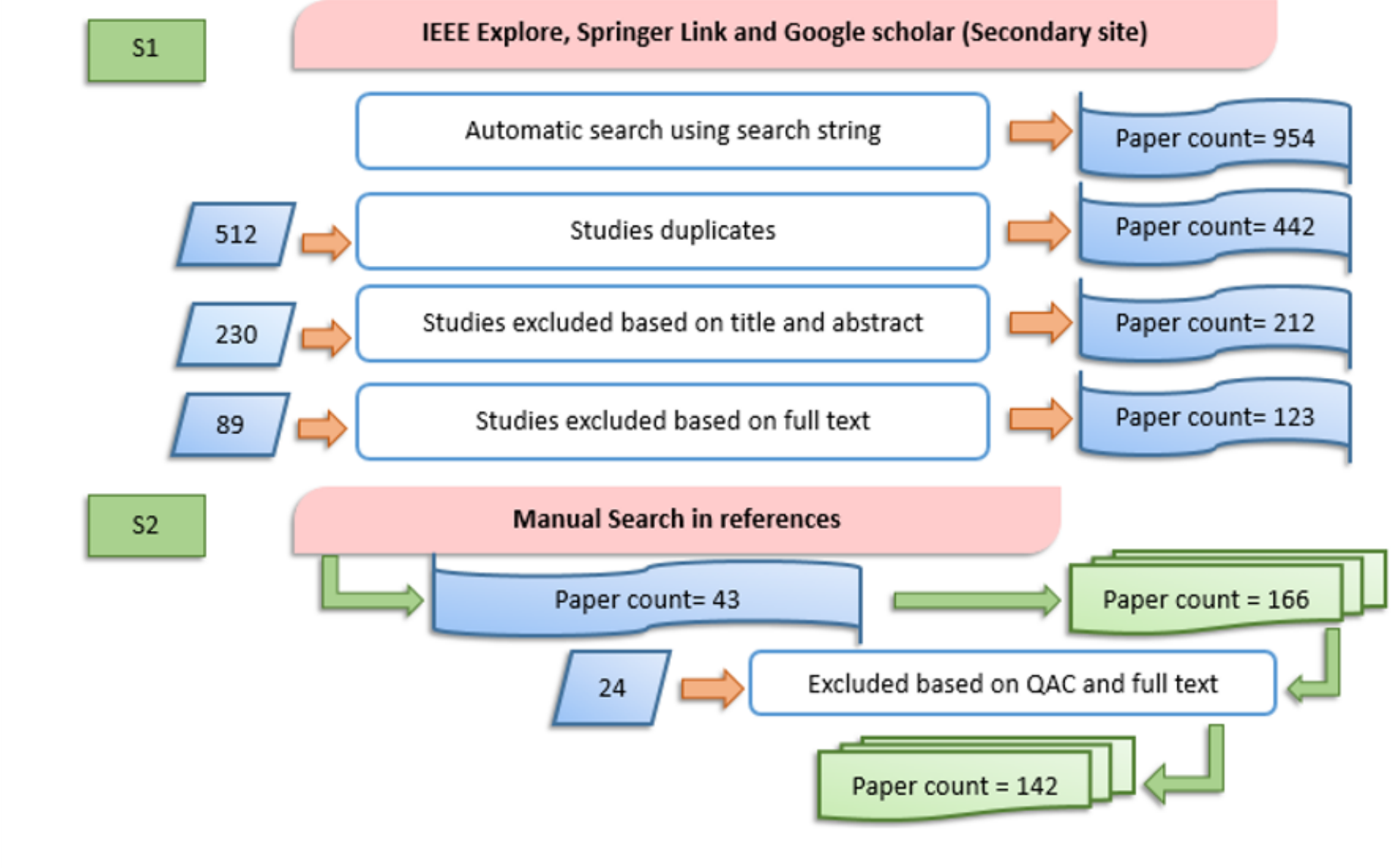}
	\caption{Compete overview of studies selection process}
	\label{fig:figure1}
\end{figure*}

Search strategy comprises of automatic and manual search, as shown in Figure \ref{fig:figure1}. An automatic search helped in identifying primary studies and to achieve a broader perspective. Therefore, we extended the review by inclusion of additional studies. As recommended by Kitchenham et al. \cite{kitchenham2010systematic}, manual search strategy was applied on the references of the studies that are identified after application of automatic search. 

For automatic search, we used standard databases which contain the most relevant research articles. These databases include IEEE Explore, ISI Web of Knowledge, Scopus—Elsevier and Springer. While there is plenty of literature available in magazine, working papers, news papers, books and blogs, we did not choose them for this review article as concepts discussed in these sources are not subjected to review process, thus their quality can't be reliably verified. 

General keywords derived from our research questions and title of study were used to search research articles. Our aim was to identify as many relevant articles as possible from main set of keywords. All possible permutations of Optical character recognition concepts were tried in the search, such as ``optical character recognition'', ``pattern recognition and  OCR'', ``pattern matching and OCR'' etc. 

Once the primary data was obtained by using search strings, data analysis phase of the obtained research papers began with the intention of considering their relevance to research questions and inclusion and exclusion criteria of study. After that, a bibliography management tool i.e. Mendeley was used for storing all related research articles to be used for referencing purpose.  Mendeley also helped us in identifying duplicate studies, because a research paper can be found in multiple databases. 

Manual search was performed with automatic search to make sure that we had not missed anything. This was achieved through forward and backward referencing. Furthermore, for data extraction all the results were imported into spreadsheet. Snowballing, which is an iterative process in which references of references are verified to identify more relevant literature, was applied on primary studies in order to extract more relevant primary studies. Set of primary studies post snowball process was then added to Mendeley.

\subsection{Study selection process} \label{SSP}

\begin{figure}[!htb]
\centering 
	\includegraphics [scale=1]{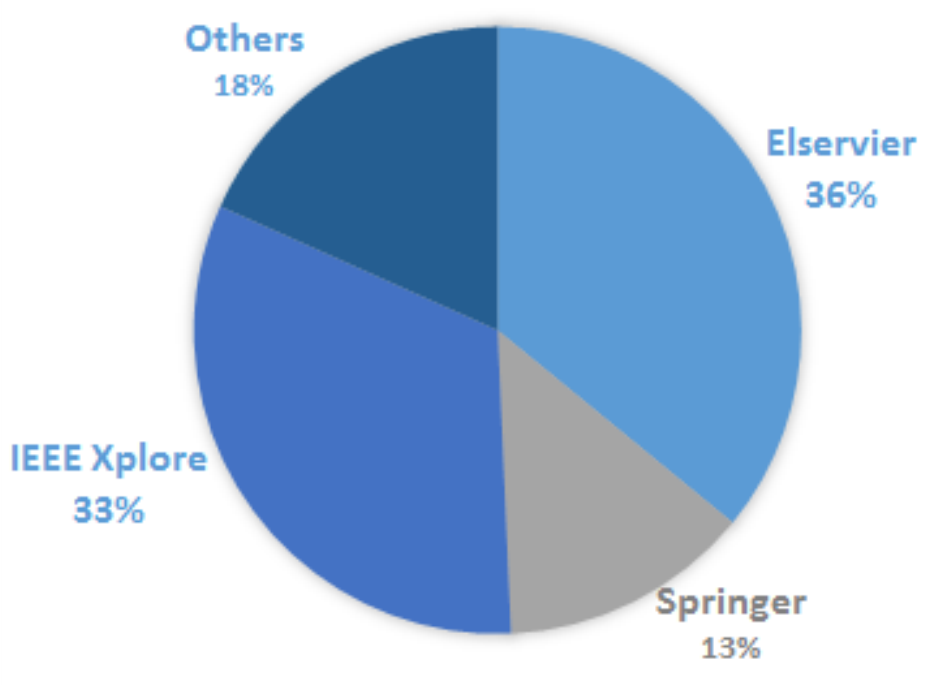}
	\caption{Distribution of sources / databases of selected studies after applying selection process}
	\label{fig:figure 2}
\end{figure}

A tollgate approach was adopted for the selection of the study \cite{nidhra2013knowledge}. Therefore, after searching keywords in all relevant databases, we extracted 954 research studies through automatic search. Majority of these 954 studies, 512 were duplicate studies and were eliminated. Inclusion and exclusion criteria based upon title, abstracts, keywords and the type of publication was applied on the remaining 442 studies. This resulted in exclusion of 230 studies and leaving 212 studies. In the next stage, the selection criteria was applied, thus further 89 studies were excluded and we were left with 123 studies.


Once we finished automatic search stage, we started manual search procedure to guarantee exhaustiveness of the search results.  We performed screening of remaining 123 studies and went through the references to check relevant research articles that could have been left search during automatic search. Manual search added 43 further studies. After adding these studies, pre-final list of 166 primary studies was obtained.

Next and final stage was to apply the quality assessment criteria (QAC) on pre-final list of 166 studies. Quality assessment criteria was applied at the end as this is the final step through which final list of studies for SLR was deduced. QAC usually identifies studies who's quality is not helpful in answering research question. After applying QAC, 24 studies were excluded and we were left with 142 primary studies. Refer Figure \ref{fig:figure1} for compete step-by-step overview of selection process.

Table \ref{Tab: Table 1} shows the distribution of the primary / selected studies among various publication sources, before and after applying above mentioned selection process. The same is also shown in Figure \ref{fig:figure 2}.

\begin{table} [!htb]
\centering

\caption{Distribution of databases of selected studies before and after applying selection process}
\begin{tabular}{ | l | c | c | }    \hline
\textbf{Source} &\textbf{Count before applying } &\textbf{Count after applying} \\
 &\textbf{selection process} &\textbf{selection process}\\ \hline

Elsevier & 207 & 51 \\ \hline
IEEE Xplore & 293& 46 \\ \hline
Springer& 273 & 19 \\ \hline
others & 182 & 26 \\ \hline
Total & 955 & 142 \\ \hline 

\end{tabular}
\label{Tab: Table 1}
\end{table}

\subsection{Quality assessment criteria}

Quality Assessment Criteria (QAC) is based on the principle to make a decision related the overall quality of selected set of studies \cite{kitchenham2010systematic}. Following criteria was used to assess the quality of selected studies. This criterion helped us to identify the strength of inferences and helped us in selecting the most relevant research studies for our research.

Quality Assessment criteria questions:

\begin {enumerate}
\item Are topics presented in research paper relevant to the objectives of this review article?
\item Does research study describes context of the research?
\item Does research article explains approach and methodology of research with clarity?
\item Is data collection procedure explained, If data collection is done in the study?
\item Is process of data analysis explained with proper examples?
\end {enumerate}

We evaluated 166 selected studies by using the above mentioned quality assessment questions in order to determine the credibility of a particular acknowledged study. These five QA schema is inspired by \cite{nidhra2013knowledge}.  The quality of study was measured depending upon the score of each QA question. Each question was assigned 2 marks and the study's quality was considered to be selected if it scored greater than or equal to 5 at the scale of 10. Thus, studies below the score of 5 were not included in the research. Following this criteria, 142 studies were finally selected for this review article (refer Figure \ref{fig:figure1} for compete overview of selection process).


\subsection{Data extraction and synthesis}
During this phase, meta data of selected studies (142) was extracted. As stated eralier, we used Mendeley and MS Excel to manage meta data of these studies. The main objective of this phase was to record the information that was obtained from the initial studies \cite{kitchenham2010systematic}.  The data containing study ID (to identify each study), study title, authors, publication year, publishing platform (conference proceedings, journals, etc.), citation count, and the study context (techniques used in the study) were extracted and recorded in an excel sheet. This data was extracted after thorough analysis of each study to identify the algorithms and techniques proposed by the researchers. This also helped us to classify the studies according to the languages on which the techniques were applied. Table \ref{Tab:Table 2} shows the fields of the data extracted from research studies.

\begin{table} [!htb]
\centering
	\caption{Extracted meta-data fields of selected studies}
	\begin{tabular}{| l | l |}
		\hline
		\textbf{Selected Features}  & \textbf{Description} \\ \hline
		Study identification number				 	& Exclusive identity for selected research article \\ \hline
		Reference														& Bibliographical Reference i.e. Authors, title, publication year etc \\ \hline
		Type of paper												& Journal, conference, workshop, symposium \\ \hline
		Language														& English, Urdu, Chinese, Arabic, Indian, Farsi / Persian \\ \hline
		Citation Count											& Number of Citations \\ \hline
		Technique 													& Feature extraction and classification techniques \\ \hline
		
	\end{tabular}
\label{Tab:Table 2}
\end{table}

\section{Statistical results from selected studies} \label{SR}
In this section, statistical results of the selected studies will be presented with respect to their publication sources, citation count status, temporal view, type of languages and type of research methodologies. 


\subsection{Publication sources overview}

\begin{figure}[!tbh]
\centering 
	\includegraphics [scale=0.8]{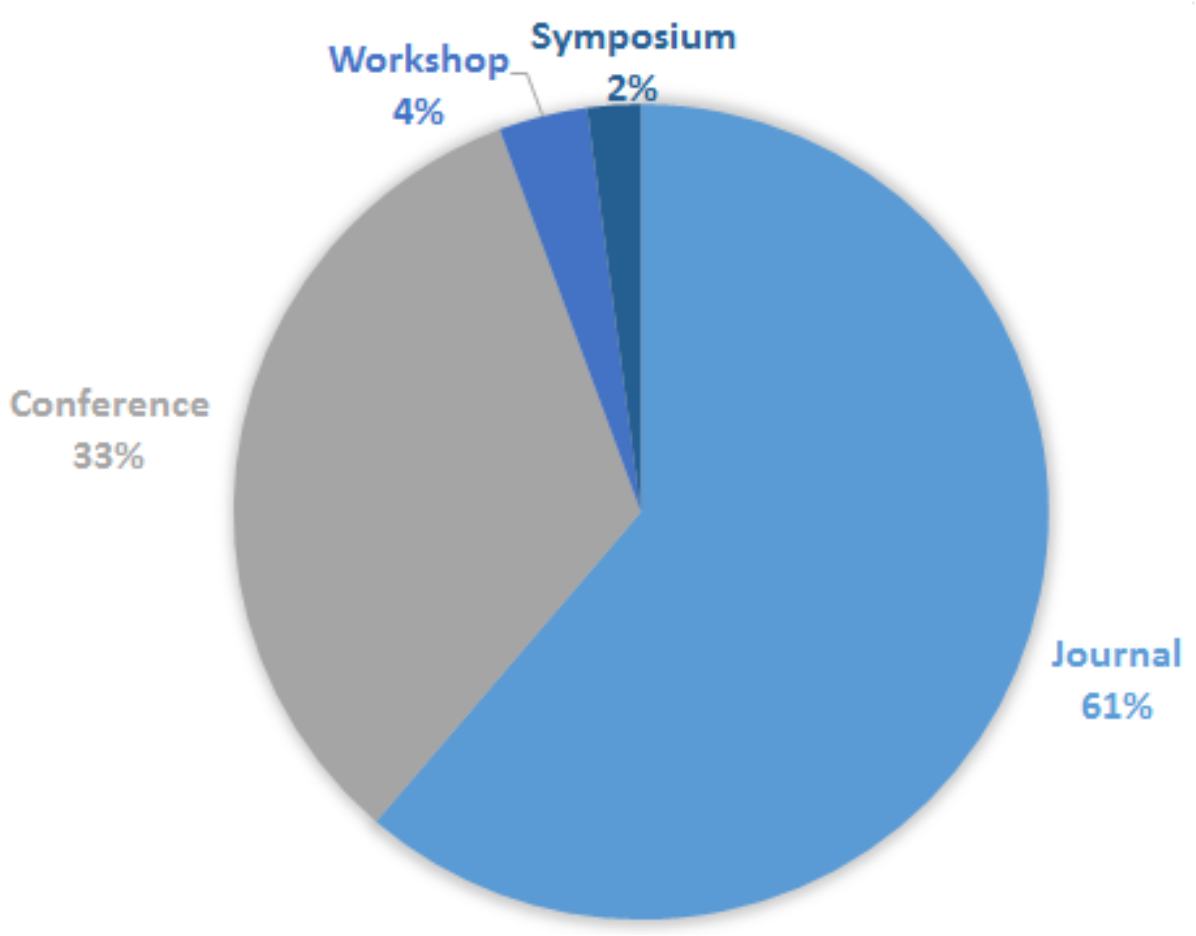}
	\caption{Study distribution per publication source}
	\label{fig:figure 3}
\end{figure}

In this review, most of the included studies are published in reputed journals and leading conferences. Therefore, considering the quality of research studies, we believe that this systematic review will be used as a reference to find latest trends and to highlight research directions for further studies in the domain of handwritten OCR. Figure \ref{fig:figure 3} shows the distribution of studies derived from different publication sources. Majority of included studies (87) were published in research journals (61\%), followed by 47 publications in conference articles (33\%). Whereas, few (5) articles were published in workshop proceedings and only 3 relevant articles were found to be presented in symposiums.

\vspace{8mm}
\subsection{Research citations} \label{ResCitation}

\begin{figure}[!htb]
	\centering
	\includegraphics [scale=0.7]{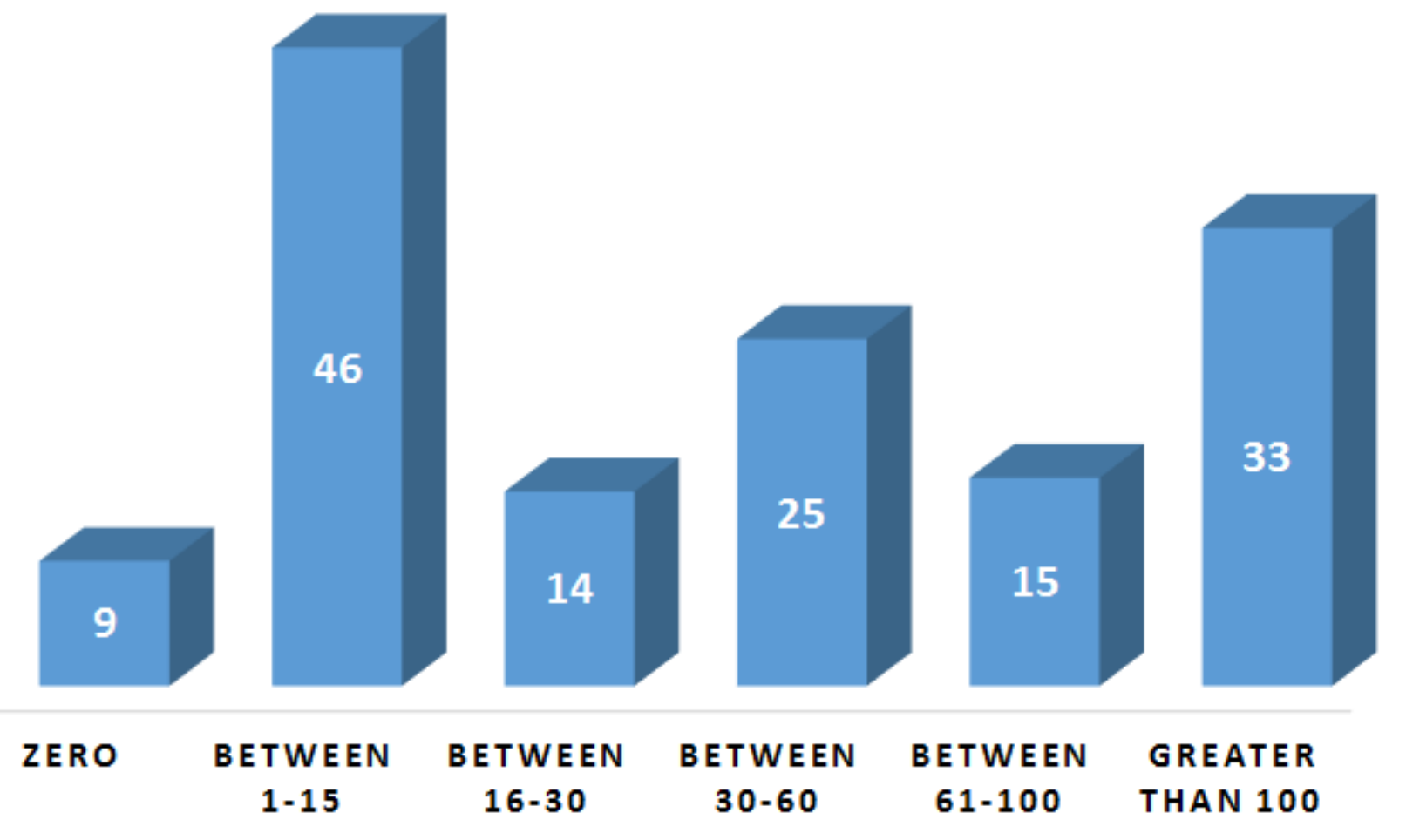}
	\caption{Citation count of selected studies. Numeric value within bar shows number of studies that have been cited $x$ times (corresponding values on the $x$-axis).}
	\label{fig:figure 4}
\end{figure}

Citation count was obtained from Google Scholar. Overall, selected studies have good citation count, which shows that quality of selected studies is worthy to be added in the review and also implies that researchers are actively working in this area of research. As presented in Figure \ref{fig:figure 4}, approximately 95\% of the selected studies have at least one citation, except few paper which are published recently in 2018. Among selected studies, 33 studies have more than 100 citations, 15 studies have been cited between 61-100 times, 25 studies were cited between 33-60 times, 14 studies were cited between 16-30 times and 46 studies were cited between 1 and 15 times. Overall, we predict that selected studies citations will increase further because research articles are constantly being published in this domain. 

Table \ref{Tab:Table 3} provides details of research publications with more than 100 citations each. These articles can be considered to have strong impact on the researchers working to build robust OCR system.

\begin{landscape}
\begin{longtable}{| p{16cm} | p{2cm}| p{1.5cm}| p{1cm}|}

	\toprule
	\label{Tab:Table 3}
\textbf{Title of Study}  &\textbf{Citations}& \textbf{Year} &\textbf{ Ref} \\ \hline	

\endhead
\multicolumn{3}{@{}l}{\ldots \textit{continued on next page}}

\endfoot
\endlastfoot

Offline handwriting recognition with multidimensional recurrent neural networks.  & 719 & 2009 & \cite{graves2009offline}\\ \hline
Handwritten numeral databases of Indian scripts and multistage recognition of mixed numerals. & 262 & 2009 & \cite{bhattacharya2009handwritten}\\ \hline
A novel connectionist system for unconstrained handwriting recognition. &1175 & 2009 &\cite{graves2009novel}\\ \hline
Markov models for offline handwriting recognition: a survey & 210 & 2009 & \cite{plotz2009markov} \\ \hline
Gujarati handwritten numeral optical character reorganization through neural network. & 148 & 2010 & \cite{desai2010gujarati}\\ \hline
Handwritten character recognition through two-stage foreground sub-sampling. & 112 & 2010 & \cite{vamvakas2010handwritten}\\ \hline
Deep, big, simple neural nets for handwritten digit recognition.  & 784 & 2010 & \cite{cirecsan2010deep}\\ \hline
Diagonal based feature extraction for handwritten character recognition system using neural network. & 175 & 2011 & \cite{pradeep2011diagonal}\\ \hline
Convolutional neural network committees for handwritten character classification. & 381 & 2011 & \cite{ciresan2011convolutional}\\ \hline
Handwritten English character recognition using neural network. & 128 & 2011 & \cite{patil2011handwritten} \\ \hline
DRAW: A recurrent neural network for image generation.  & 995 & 2015 & \cite{gregor2015draw}\\ \hline
Online and off-line handwriting recognition: a comprehensive survey.   & 2909 & 2000 & \cite{plamondon2000online}\\ \hline
Template-based online character recognition.   & 173 & 2001 & \cite{connell2001template}\\ \hline
An overview of character recognition focused on off-line handwriting. & 589 & 2001 & \cite{arica2001overview} \\ \hline
IFN/ENIT-database of handwritten Arabic words.& 477 & 2002 &\cite{pechwitz2002ifn}\\ \hline
Off-line Arabic character recognition–a review.& 236 & 2002 &\cite{khorsheed2002off}\\ \hline
A class-modular feedforward neural network for handwriting recognition. & 131 & 2002 &\cite{oh2002class}\\ \hline
Individuality of handwriting. & 552 & 2002 &\cite{srihari2002individuality}\\ \hline
HMM based approach for handwritten Arabic word recognition using the IFN/ENIT-database. & 172 & 2003 & \cite{pechwitz2003hmm}\\ \hline
Handwritten digit recognition: benchmarking of state-of-the-art techniques. & 573 & 2003 & \cite{liu2003handwritten}\\ \hline
Indian script character recognition: a survey. & 540 & 2004 & \cite{pal2004indian}\\ \hline
Online recognition of Chinese characters: the state-of-the-art. & 362 & 2004 &\cite{liu2004online}\\ \hline
A study on the use of 8-directional features for online handwritten Chinese character recognition. & 145 & 2005 & \cite{bai2005study}\\ \hline
Offline Arabic handwriting recognition: a survey. & 551 & 2006 & \cite{lorigo2006offline}\\ \hline
Recognition of off-line handwritten devnagari characters using quadratic classifier. & 168 & 2006 & \cite{sharma2006recognition}\\ \hline
Connectionist temporal classification: labeling unsegmented sequence data with RNN. & 1404 & 2006 & \cite{graves2006connectionist}\\ \hline
Text-independent writer identification and verification on offline arabic handwriting. & 128 & 2007 & \cite{bulacu2007text}\\ \hline
A novel approach to on-line handwriting recognition based on bidirectional LSTM networks. & 172 & 2007 & \cite{liwicki2007novel}\\ \hline
Fuzzy model based recognition of handwritten numerals. & 148 & 2007 & \cite{hanmandlu2007fuzzy}\\ \hline
Introducing a very large dataset of handwritten Farsi digits and a study on their varieties. & 155 & 2007 & \cite{khosravi2007introducing}\\ \hline
Unconstrained on-line handwriting recognition with recurrent neural networks & 207 & 2007 & \cite{graves2008unconstrained}\\ \hline
ICDAR 2013 Chinese handwriting recognition competition. & 177 & 2013 & \cite{yin2013icdar}\\ \hline

Automatic segmentation of the IAM off-line database for handwritten English text. &101 &2002 &\cite{zimmermann2002automatic}\\ \hline
\caption{Research publications with more than 100 citations}
\end{longtable}
\end{landscape}

\subsection{Temporal view}

\begin{figure}[!htb]
\centering
	\includegraphics[scale=0.75]{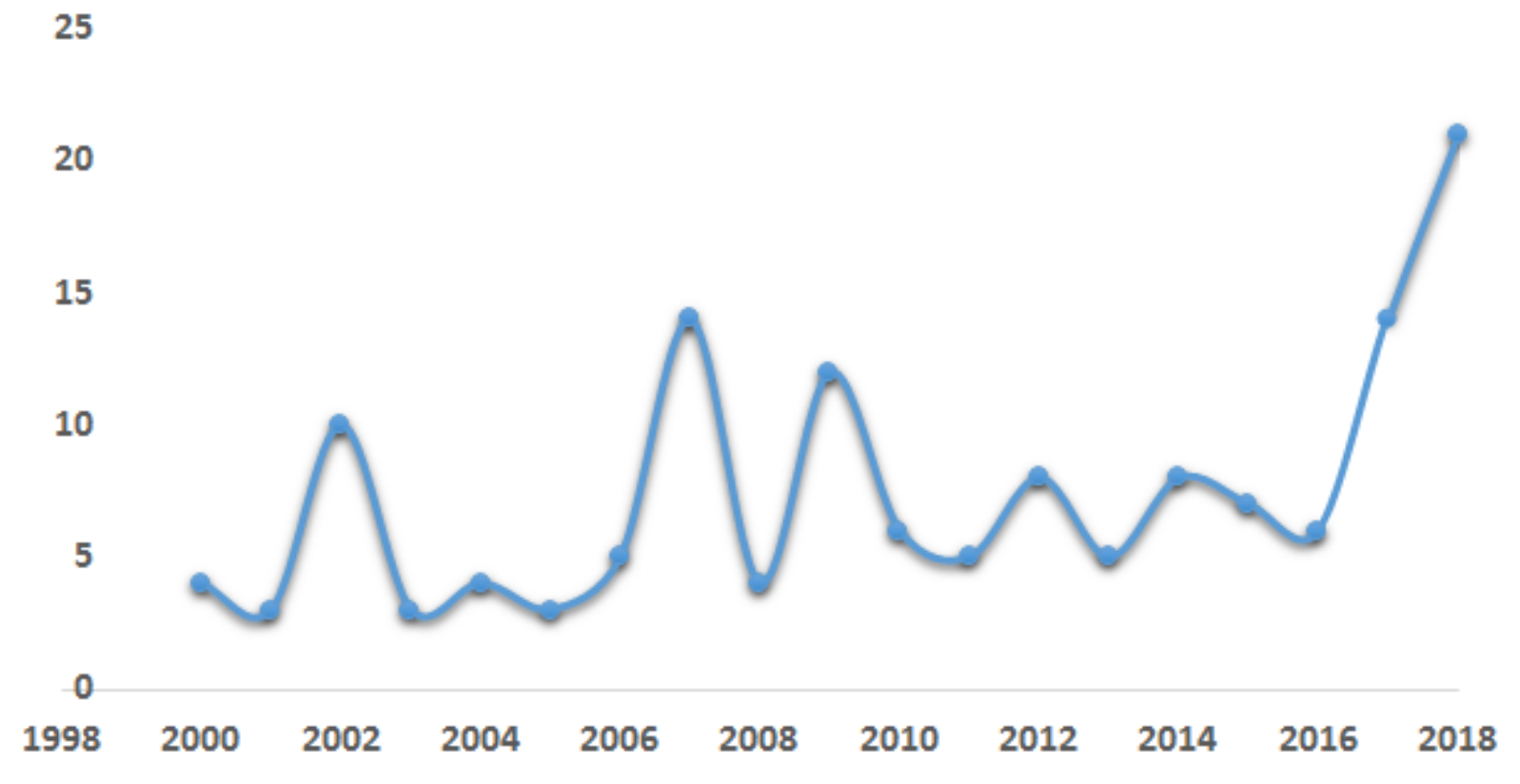}
	\caption{Publications Over The Years. On the $y$-axis is the number of publications. }
	\label{fig:figure 5}
	
\end{figure}

The distribution of number of studies over the period under study (2000 - 2018) can be seen in Figure \ref{fig:figure 5}. According to the reference figure, it can be noticed that there is a variation in the publication count through these years. Statistics show sudden increase in number of publications in the domain of a handwritten character recognition in the years 2002, 2007 and 2009. The number of publications remained steady in the remaining years of 2000s. After 2010 there is again steady increase in the number publications i.e. 59 publications in last 8 years.  Year 2017 and 2018 have seen a steady rise in number of publications. This is conceivably not surprising, since the concept of a handwritten character recognition is catching interest of more researcher because of the advancement of the research work in the fields of deep learning and computer vision. We believe that application areas of Handwritten OCRs will further increase in the coming years.

\subsection{Language specific research} \label{LSR}

\begin{figure}[!htb]
	\centering
		\includegraphics [scale=0.75]{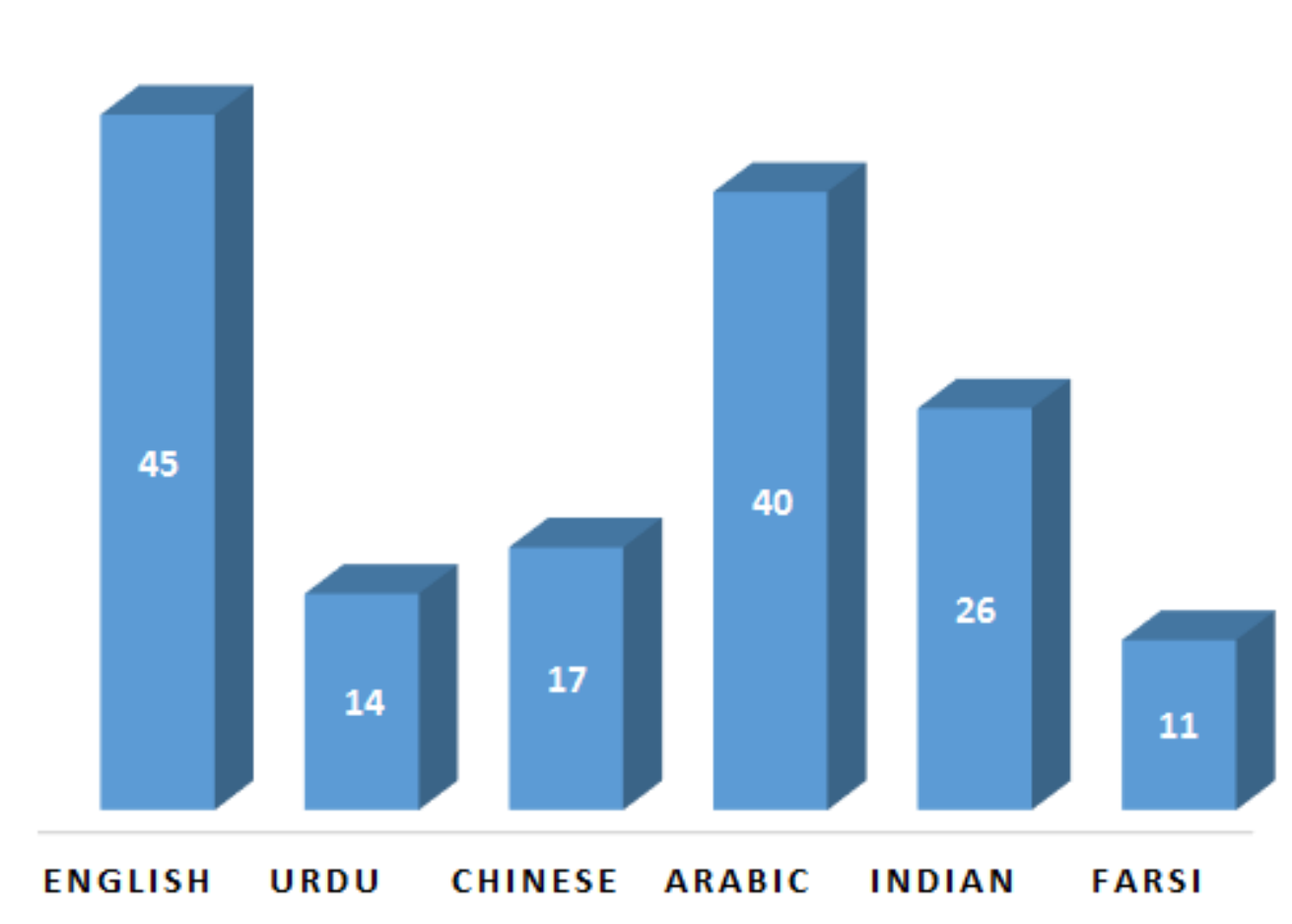}
	\caption{Number of selected studies with respect to investigated language. Numeric value within bar shows number of  selected studies for the given language.}
	\label{fig:figure 6}
\end{figure}

The distributions / number of selected studies with respect to investigated scripting languages are shown in the Figure \ref{fig:figure 6}. Total number of selected studies are 142 and out of these 142 studies, English language has the highest contribution of 45 studies in the domain of character recognition, 40 studies related to Arabic language, 26 studies are on the Indian scripts, 17 on Chinese language, 14 on Urdu language, while 11 studies were conducted on Persian language. Some of the selected articles discussed multiple languages. 

\begin{figure}[!htb]
	\centering
		\includegraphics  [scale=0.7]{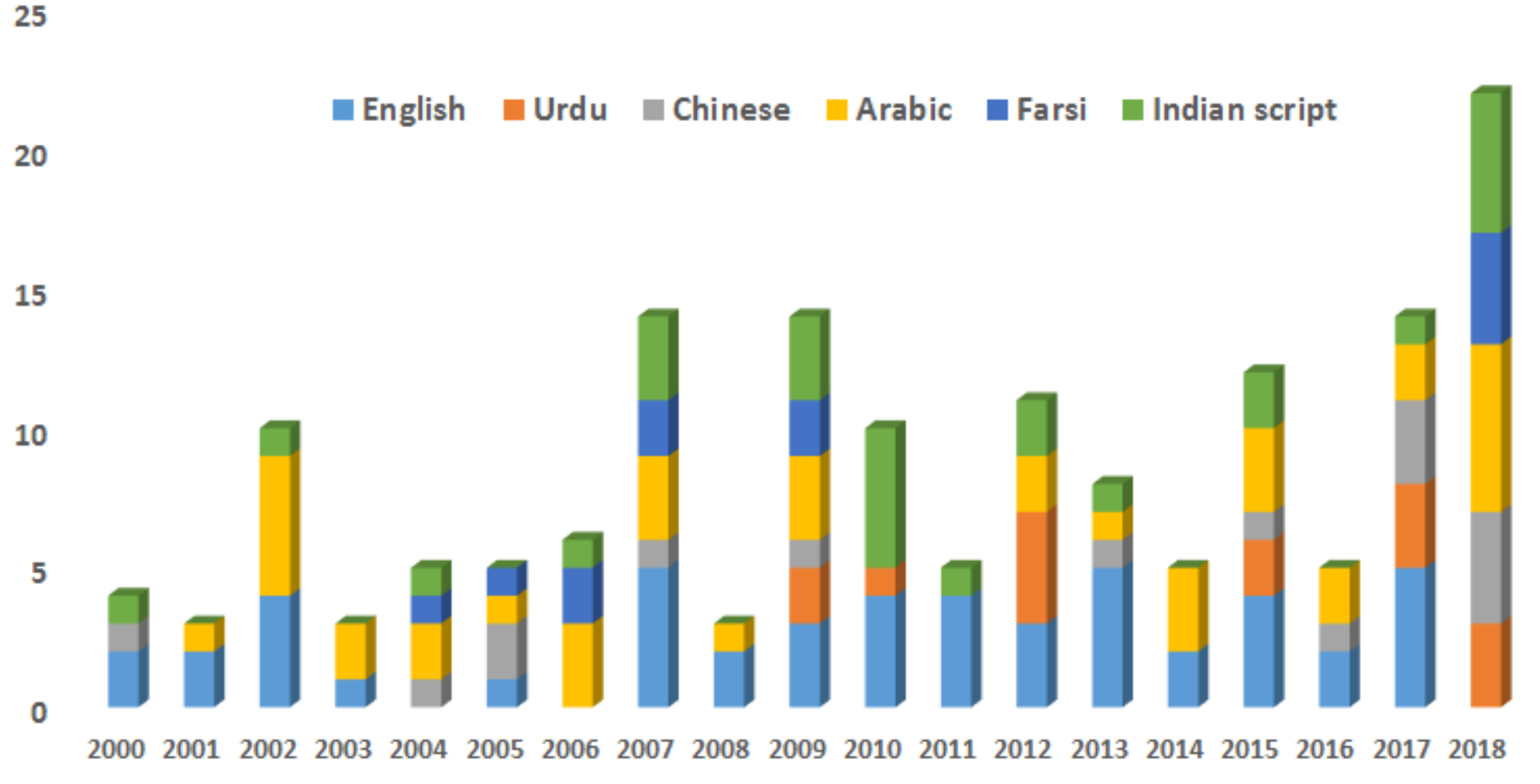}
	\caption{Selected studies count each year with respect to specific language. $y$-axis shows the number of selected studies. Specific color within each bar represents specific language as shown in the legend.}
	\label{fig:figure 7}
\end{figure}

Figure \ref{fig:figure 7} represents publications count each year with respect to language. Reference figure shows compiled temporal view of handwritten OCR researches done in different languages throughout the mentioned era of 2000-2018, in this time period there are certain research articles that covers more than one language of handwritten OCR. 


\section{Research questions} \label{RQ}
Research questions play an important role in systematic literature review, because these questions determine the search queries and keywords that will be used to explore research publications. As discussed above, we chose research questions which not only help  seasoned researchers but also to researchers entering in the domain of optical character recognition to understand where the research in this field stands as of today. This review article answers research questions presented in Table \ref{Tab:RQs}. Reference table also presents motivation for each research question.  

\begin{table} [!htb]
\centering
	
	\caption{Research questions and motivation}
	\begin{tabular}{| p{8cm} |  p{8cm} |}
		\hline
		\textbf{Research question}  & \textbf{Motivation} \\ \hline
		
		What different feature extraction and classifications methods are used for handwritten OCR?  & To identify trends in used feature extractors and machine learning techniques over almost two decades.  \\ \hline


		What different datasets / databases are  available for research purpose?  & Availability of a dataset with enough data is always fundamental requirement for buidling OCR system \cite{KHAN201961}  \\ \hline

		What major languages are investigated?   & To highlight which languages have usually been investigated. Thus identifying languages which needs more research attention. \\ \hline
		
		What are the new research domains in the area of OCR?   & To provide guidance for new research projects.  \\ \hline
		
	\end{tabular}
\label{Tab:RQs}
\end{table}


\section{Classification methods of handwritten OCR} \label{classi}

In handwritten OCR an algorithm is trained on a known dataset and it discovers how to accurately categorize / classify the alphabets and digits. Classification is a process to learn model on a given input data and map or label it to predefined category or classes \cite{Mitchell}. In this section we have discussed most prevalent classification techniques in OCR research studies beginning from 2000 till 2018.

\subsection{Artificial Neural Networks (ANN)}

\begin{figure}[!htb]
	\centering
	\includegraphics [scale=0.27]{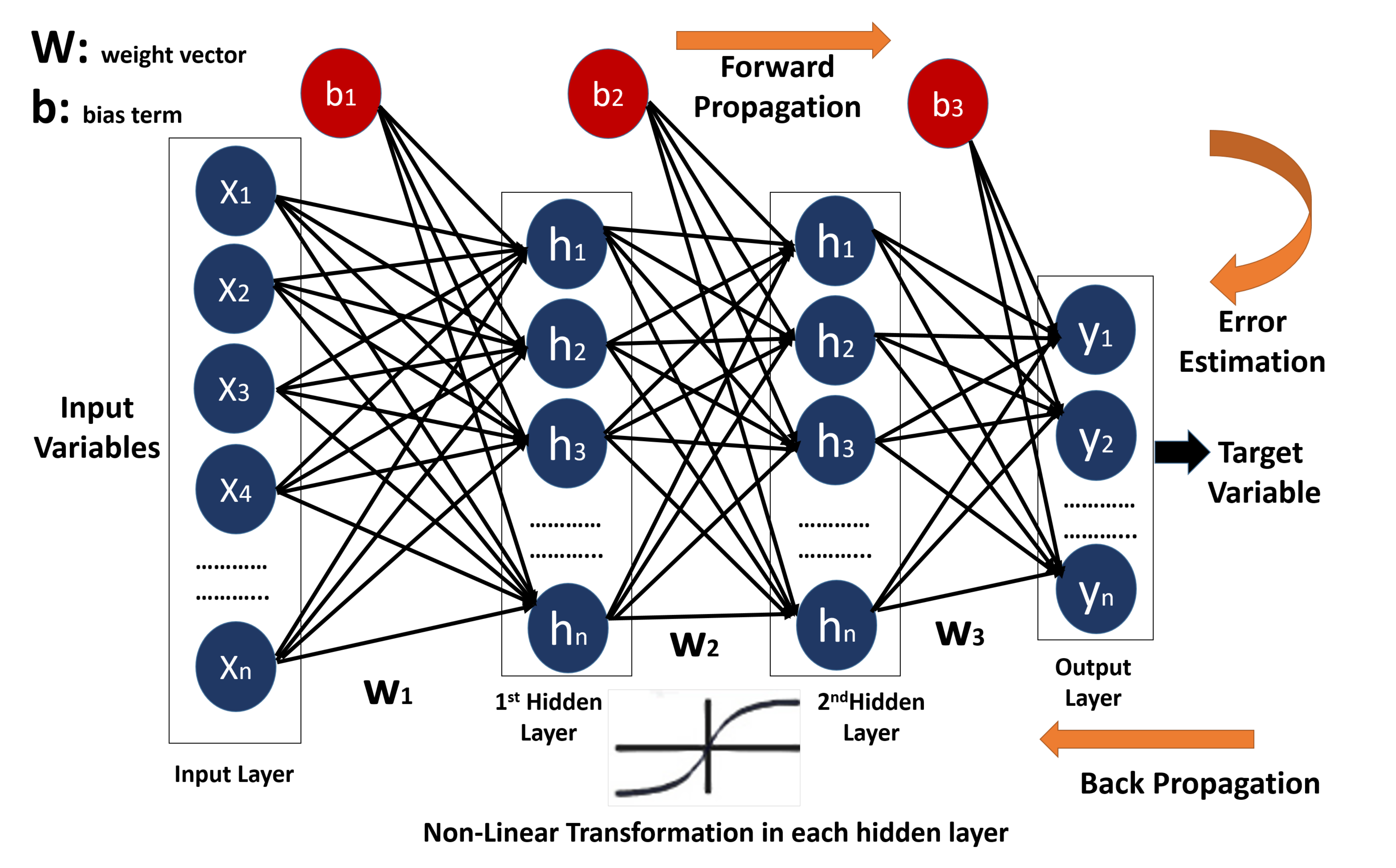}
	\caption {An architecture of Multilayer Perceptron (MLP) \cite{RizHam}}
	\label{fig:figure 12}
\end{figure}

Biological neuron inspired architecture, Artificial Neural Networks (ANN) consists of numerous processing units called neurons \cite{NNBook}. These processing elements (neurons) work together to model given input data and map it to predefined class or label \cite{vithlani2015study}. The main unit in neural networks is nodes (neuron). Weights associated with each node are adjusted to reduce the squared error on training samples in a supervised learning environment (training on labeled samples / data). Figure \ref{fig:figure 12} presents pictorial representation of Multi Layer Perceptron (MLP) that consists of three layers i.e. (input, hidden and output).


Feed forward networks / Multi Layer Perceptron (MLP) achieved renewed interest of research community in mid 1980s as by that time ``Hopfield network'' provided the way to understand human memory and calculate state of a neuron \cite{485891}. Initially, computational complexity of finding weights associated with neurons hindered application of neural networks. With the advent of deep (many layers) neural architectures i.e.  Recurrent Neural Network (RNN) and Convolutional Neural Networks (CNN), neural networks has established it self as one of the best classification technique for recognition tasks including OCR \cite{nawaz2003approach,srihari2007offline,pradeep2012neural,singh2011feature}. Refer Sections \ref{future} and \ref{fworks} for current and future research trends.

The early implementation of MLP in handwritten OCR was done by shamsher et al. \cite{shamsher2007ocr} on Urdu language. The researchers proposed feed forward neural network algorithm of MLP (Multi Layer Preceptrons) \cite{al2009handwriting}. Liu et al. \cite{liu2009new} used MLP on Farsi and Bangla numerals. One hidden layer was used with the connecting weights estimated by the error back-propagation (BP) algorithm that minimized the squared error criterion. On the other hand, Ciresan et al \cite{cirecsan2010deep} trained five MLPs with two to nine hidden layers and varying numbers of hidden units for the recognition of English numerals.

Recently, Convolutional Neural Network (CNN) has reported great success in character recognition task \cite{liu2005classification}. Convolutional neural network has been widely used for classification and recognition of almost all the languages that have been reviewed for this systematic literature review \cite{boufenar2018investigation,sokar2018generic,lin2018chinese,yang2018recognition,alizadehashraf2017persian,ghasemi2018persian}.

\subsection{Kernel methods}

A number of powerful kernel-based learning models, e.g. Support Vector Machines (SVMs), Kernel Fisher Discriminant Analysis (KFDA) and Kernel Principal Component Analysis (KPCA) have shown practical relevance for classification problems. For instance, in the context of optical pattern, text categorization, time-series prediction these models have significant relevance.

In support vector machine, kernel performs mapping of feature vectors into a higher dimensional feature space in order to find a hyperplane, which is linearly separates classes by as much margin as possible. Given a training set of labeled examples \{ ($x_i, y_i$) , $i$ = 1 \dots $l$ \} where $x_i$ $\in$ $\Re^n$ and $y_i$ $\in$ \{-1, 1\}, a new test example $x$ is classified by the following function:

\begin{equation}
f(x)=sgn(\sum\limits_{i=1}^{l} \alpha_i y_i K(x_i,x)+b)
\end{equation}

where:
\begin {enumerate}
\item $K(.,.)$ is a kernel function
\item $b$ is the threshold parameter of the hyperplane
\item $\alpha_i$ are Langrange multipliers of a dual optimization problem that describe the separating hyperplane
\end {enumerate}

Before popularization of deep learning methodology, SVM was one of the most robust technique for handwritten digit recognition, image classification, face detection, object detection, and text classification \cite{boukharouba2017novel}. Kernel Fisher Discriminant Analysis (KFDA) and Kernel Principal Component Analysis (KPCA) are also some of the most significant kernel methods being used in offline handwritten character recognition system \cite{verma2012survey}.

Boukharouba et al. \cite{boukharouba2017novel,yang2005discrimination} used SVM for recognition of Urdu and Arabic handwritten digits. SVMs have also been successfully implement in image classification and affect recognition \cite{yang2009linear, KHAN20131159}, text classification \cite{haddoud2016combining} and face and object detection\cite{ning2016object,tao2016robust}. 


\subsection{Statistical methods}
Statistical classifiers can be parametric and non-parametric. Parametric classifiers have fixed (finite) number of parameters and their complexity is not a function of size of input data. Parametric classifiers are generally fast in learning concept and can even work with small training set. Example of parametric classifiers are Logistic Regression (LR), Linear Discriminant Analysis (LDA), Hidden Markov Model (HMM) etc.

On the other hand, non-parametric classifiers are more flexible in learning concepts but usually grow in complexity with the size of input data.  $K$ Nearest Neighbor ($K$NN), Decision Trees (DT) are examples of non-parametric techniques as as their number of parameters grows with the size of the training set.

\subsubsection{Non-parametric statistical methods}
One of the most used and easy to train statistical model for classification is $k$ nearest neighbor ($k$NN) \cite{liu2003handwritten,akbari2018novel,chandio2018character}. It is a non-parametric statistical method, which is widely used in optical character recognition. Non-parametric recognition does not involve a-priori information about the data.

$k$NN finds number of training samples closest to new example based on target function. Based upon the value of targeted function, it infers the value of output class. The probability of an unknown sample $q$ belonging to class $y$ can be calculated as follows: 


\begin{equation}\label{KNN}
p (y\mid q) = \frac {\sum_ {k \epsilon K} W_{k} .1_{(k_{y}=y)}}{\sum_ {k \epsilon K} W_{k}}
\end{equation}

\begin{equation}\label{KNNS}
W_{k}= \frac {1} {d(k,q)}
\end{equation}

$where$;

\begin {enumerate}
\item $K$ is the set of nearest neighbors
\item $k_{y}$ the class of $k$
\item $d(k,q)$ the Euclidean distance of $k$ from $q$, respectively.
\end {enumerate}



Researchers have been found to use  $k$NN for over a decade now and they believe that this algorithm achieves relatively good performance for character recognition in their experiments performed on different datasets \cite{pradeep2012neural, lorigo2006offline, chandio2018character, kumar2018improved}. 

$k$NN classifies object / ROI based on majority vote of its neighbors (class) as it assigns class most prevalent among its $k$ nearest neighbors. If k = 1, then the object is simply assigned to class of that single nearest neighbor \cite{vithlani2015study}.

\subsubsection{Parametric statistical methods}

As mentioned above parametric techniques models concepts using fixed (finite) number of parameters as they assume sample population / training data can be modeled by a probability distribution that has a fixed set of parameters. In OCR research studies, generally characters are classified according to some decision rules such as maximum likelihood or Bayes method once parameters of model are learned \cite{arica2001overview}.




Hidden Markov Model (HMM) was one the most frequently used parametric statistical method earlier in 2000.

HMM models system / data that is assumed to be Markov process with hidden states, where in Markov process probability of one states only depends on previous state \cite{arica2001overview}. It was first used in speech recognition during 1990s before researchers started using it in recognition of optical characters \cite{alma2002recognition, alma2004off, cheriet2008visual}. It is believed that HMM provides better results even when availability of lexicons is limited \cite{pechwitz2003hmm}.

\subsection{Template matching techniques}

\begin{figure} [!htb]
	\centering
	\includegraphics [scale=0.7]{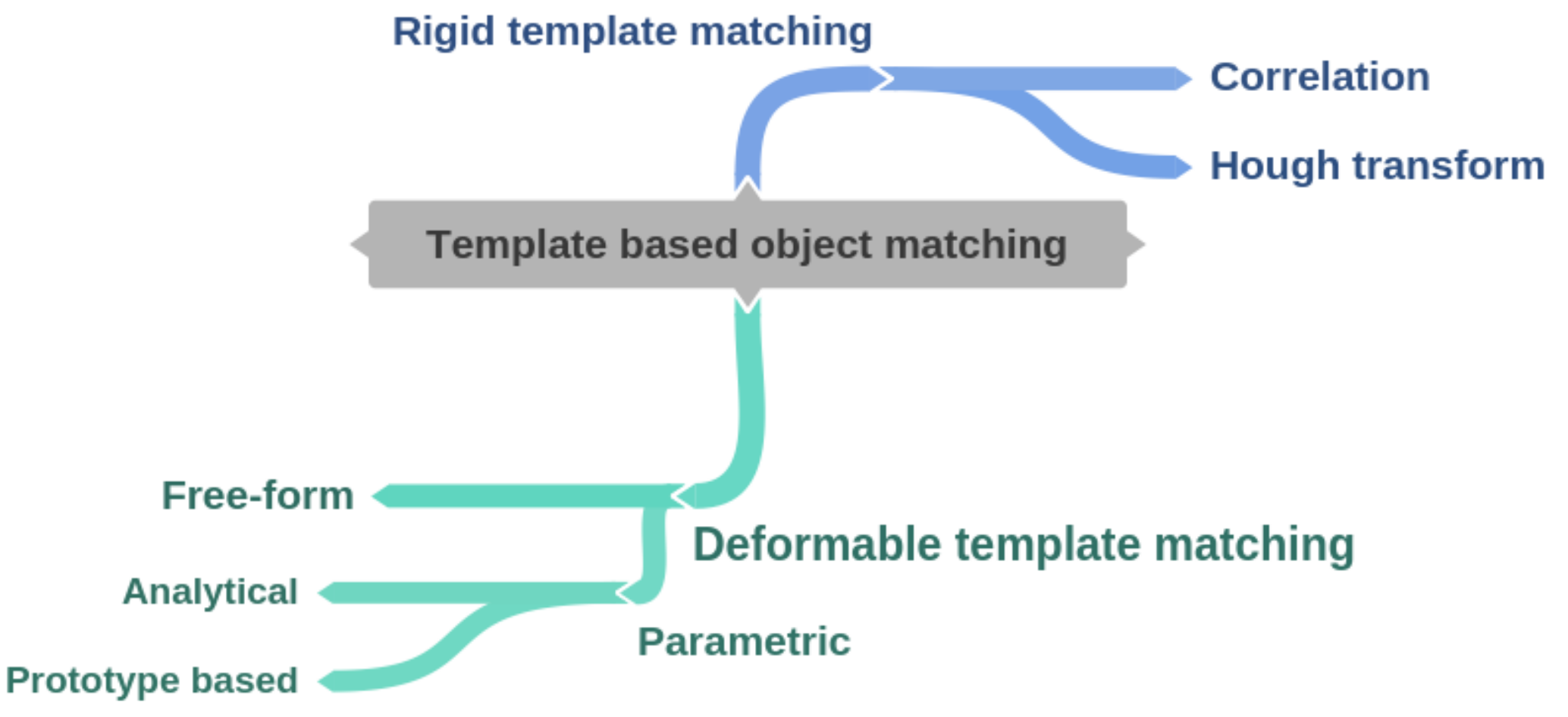}
	\caption{An overview of template matching techniques}
	\label{fig:figure 8}
\end{figure}

As the names suggests, template matching is an approach in which images (small part of an image) is matched with certain predefined template. Usually template matching techniques employ sliding sliding window approach in which template image or feature are slided on the image to determine similarity between the two.  Based on used similarity (or distance) metric classification of different objects are obtained \cite{pal2012handwriting}. 

In OCR, template matching technique is used to classify character after matching it with predefined template(s) \cite{sahu2013offline}. In literature, different distance (similarity) metrics are used, most common ones are Euclidean distance, city block distance, cross correlation, normalized correlation etc. 

In template matching, either template matching technique employs rigid shape matching algorithm or deformable shape matching algorithm. Thus, creating different family of template matching. Taxonomy of template matching techniques is presented in Figure \ref{fig:figure 8}.



\begin{figure} [!htb]
	\centering
	\includegraphics [height=5cm, width=8cm]{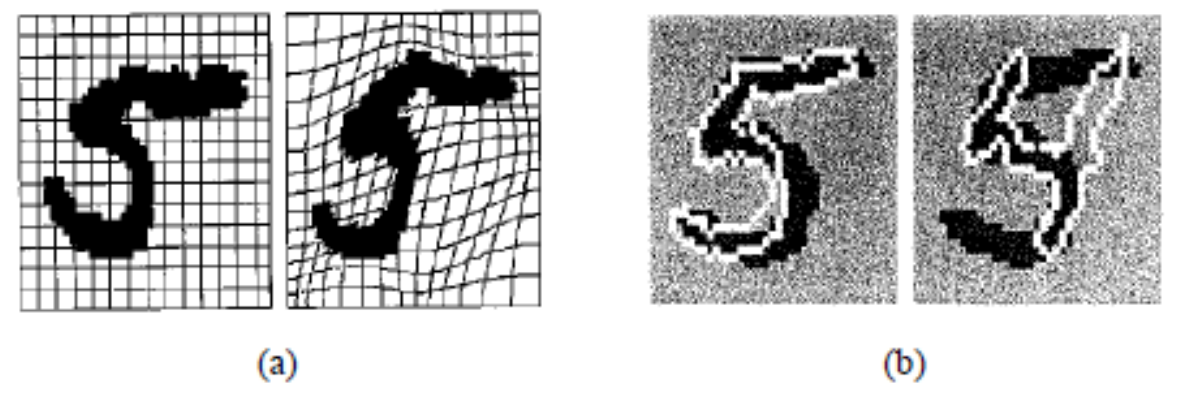}
	\caption{(a) Digit Deformations (b) Deformed Template superimposed on target image \cite{arica2001overview}}
	\label{fig:figure 9}
\end{figure}

One of the most applicable approach for character recognition is deformable template matching (refer Figure \ref{fig:figure 9}) as different writers can write character by deforming them in particular way specific to writer. In this approach, deformed image is used to compare it with database of known images. Thus, matching / classification is performed with deformed shapes as specific writer could have deformed character in a particular way \cite{arica2001overview}. Deformable template matching is further divided into parametric and free form matching. Prototype matching, which is sub-class of parametric deformable matching, matching of done based on stored prototype (deformed) \cite{tiwari2012novel}.

Apart from deformable template matching approach, second sub-class of template matching is rigid template matching. As the name suggests, rigid template matching does not take into account shape deformations. This approach usually works with features extraction / matching of image with template. One of the most common approach used in OCR to extract shape features is Hough transform, like Arabic \cite{4376996} and Chinese \cite{Li1995}.

Second sub-class of rigid template matching is correlation based matching. In this technique, initially image similarity is calculated and based on similarity features from specific regions are extracted and compared \cite{arica2001overview, chaudhuri2010some}. 





\subsection{Structural pattern recognition}
\begin{figure} [!htb]
	\centering
	\includegraphics [scale=0.75]{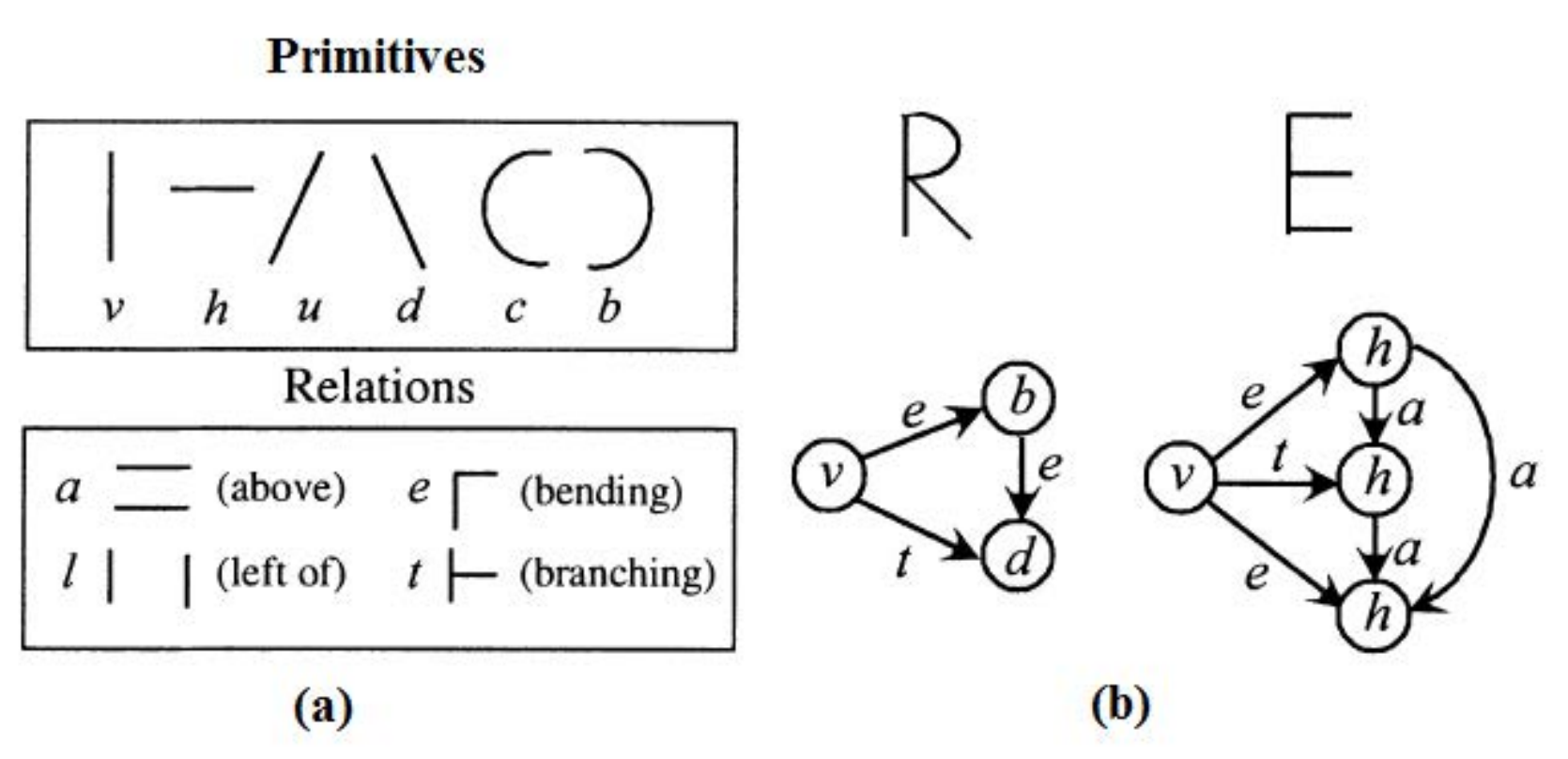}
	\caption{(a) Primitive and relations (b) Directed graph for capital letter R and E \cite{Marquesde2001}}
	\label{fig:graph}
\end{figure}

Another classification technique that was used by OCR research community before the popularization of kernel methods and neural networks / deep learning  approach was structural pattern recognition. Structural pattern recognition aims to classify objects based on relationship between its pattern structures and usually structures are extracted using pattern primitives (refer Figure \ref{fig:graph} for an example of pattern primitives) i.e. edge, contours, connected component geometry etc . One of such image primitive that has been used in OCR is Chain Code Histogram (CCH) \cite{bookCCH, LIU2004265}. CCH effectively describes image / character boundary / curve, thus helping in classify character \cite{boukharouba2017novel, vithlani2015study}. Prerequisite condition to apply CCH for OCR is that image should be in binary format and boundaries should be well define. Generally, for handwritten character recognition this condition makes CCH difficult to use. Thus, different research studies and publicly available datasets use / provide binarized images \cite{akbari2018novel}.

In research studies of OCR, structural models can be further sub-divided on the basis of context of structure i.e. graphical methods and grammar based methods . Both of these models are presented in next two sub-sections.



\subsubsection{Graphical methods}

A graph ($G$) is a way to mathematically describe relation between connected objects and is represented by ordered pair of nodes ($N$) and edges ($E$). Generally for OCR, $E$ represents arc of writing stroke connecting $N$. The particular arrangement  of $N$ and $E$ define characters / digits / alphabets. Trees (undirected graph, where direction of connection is not defined), directed graphs (where direction of edge to node is well defined) are used in different research studies to represent characters mathematically \cite{rohtua, ALVARO201458}. 

As mentioned above, writing structural components are extracted using pattern primitives i.e. edge, contours, connected component geometry etc. Relation between these structures can be defined mathematically using graphs (refer Figure \ref{fig:graph} for an example showing how letter ``R'' and ``E'' can be modeled using graph theory). Then considering specific graph architecture different structures can be classified using graph similarity measure i.e. similarity flooding algorithm \cite{994702}, SimRank algorithm \cite{Jeh:2002}, Graph similarity scoring \cite{ZAGER200886} and vertex similarity method \cite{Leicht2006}. In one study \cite{8263187}, graph distance is used to segment overlapping and joined characters as well.





\subsubsection{Grammar based methods}

In graph theory, syntactic analysis is also used to find similarities in graph structural primitives using concept of grammar \cite{grammar}. Benefit of using grammar concepts in finding similarity in graphs comes from the fact that this area is well researched and techniques are well developed.  There are different types of grammar based on restriction rules, for example unrestricted grammar, context-free grammar, context-sensitive grammar and regular grammar. Explanation of these grammar and corresponding applied restrictions are out scope of this survey article. 

In OCR literature, usually strings and trees are used to represent models based on grammar.  With well defined grammar, string is produced that then can be robustly classified to recognize character. Tree structure can also models hierarchical relations between structural primitives \cite{pal2012handwriting}. Trees can also be classified by analyzing grammmar that defines the tree, thus classifying specific character \cite{Chaudhuri2017}.


\section{Datasets}\label{dataset}

Generally, for evaluating and benchmarking different OCR algorithms, standardized databases are needed / used to enable a meaningful comparison \cite{KHAN201961}. Availability of a dataset containing enough amount of data for training and testing purpose is always fundamental requirement for a quality research \cite{hussain2015comprehensive, Khan2011b}. Research in the domain of optical character recognition mainly revolves around six different languages namely, English, Arabic, Indian, Chinese, Urdu and Persian / Farsi script. Thus, there are publicly available datasets for these languages such as MNIST, CEDAR, CENPARMI, PE92, UCOM, HCL2000 etc. 


Following subsections presents an overview of most used datasets for above mentioned languages. 

\subsection{CEDAR}

\begin{figure}[!htb]
	\centering
	\includegraphics [scale=0.15]{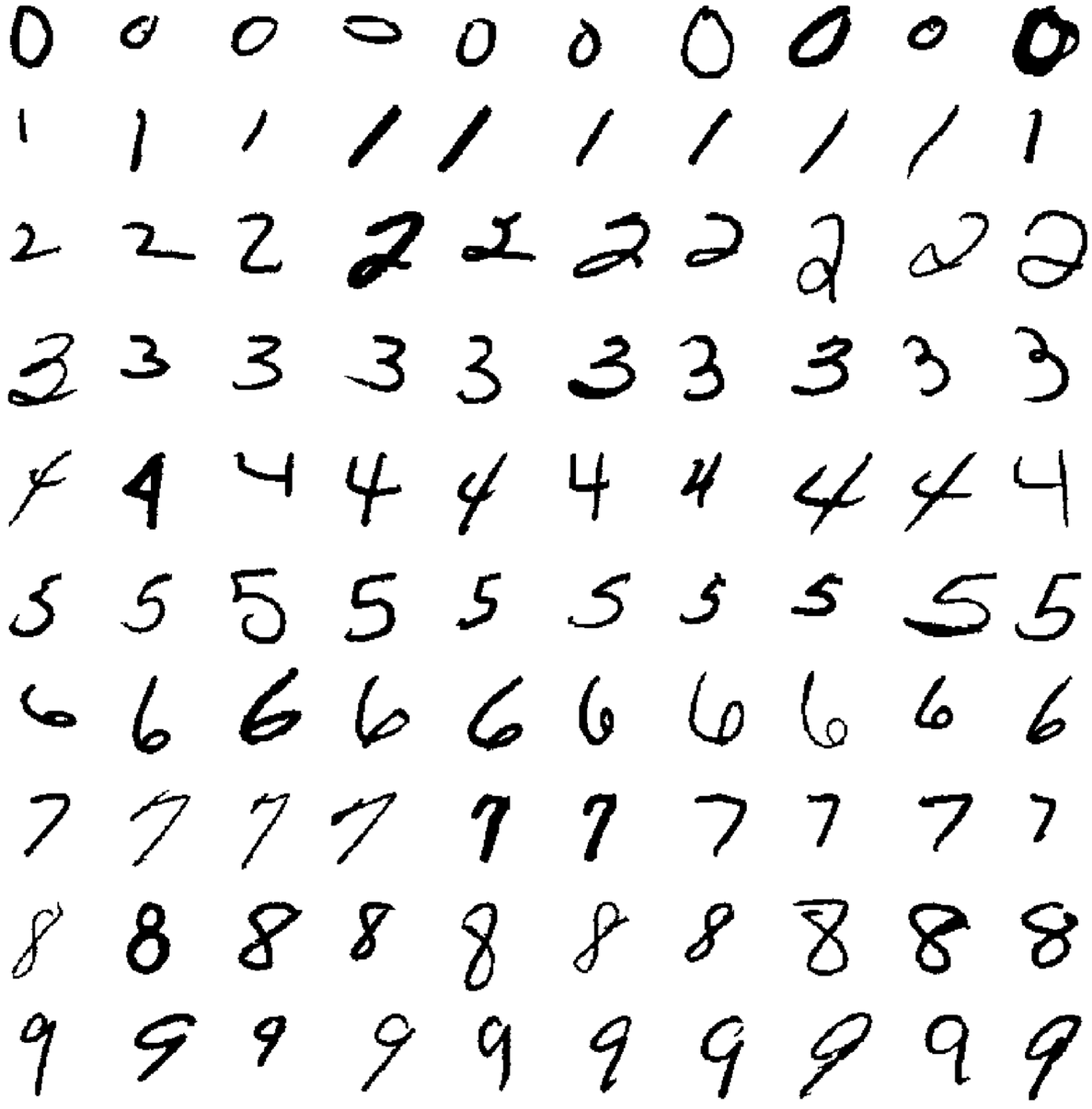}
	\caption{Sample image from CEDAR Dataset \cite{liu2003handwritten}}
	\label{fig:figure 16}
\end{figure} 

This legacy dataset, CEDAR, was developed by the researchers at University of Buffalo in 2002 and is considered among the first few large databases of handwritten characters \cite{srihari2002individuality}. In CEDAR the images were scanned at 300 dpi. Example character images from CEDAR database are shown in Figure \ref{fig:figure 16}.


\subsection{MNIST}

\begin{figure} [!htb]
	\centering
	\includegraphics [scale=0.6]{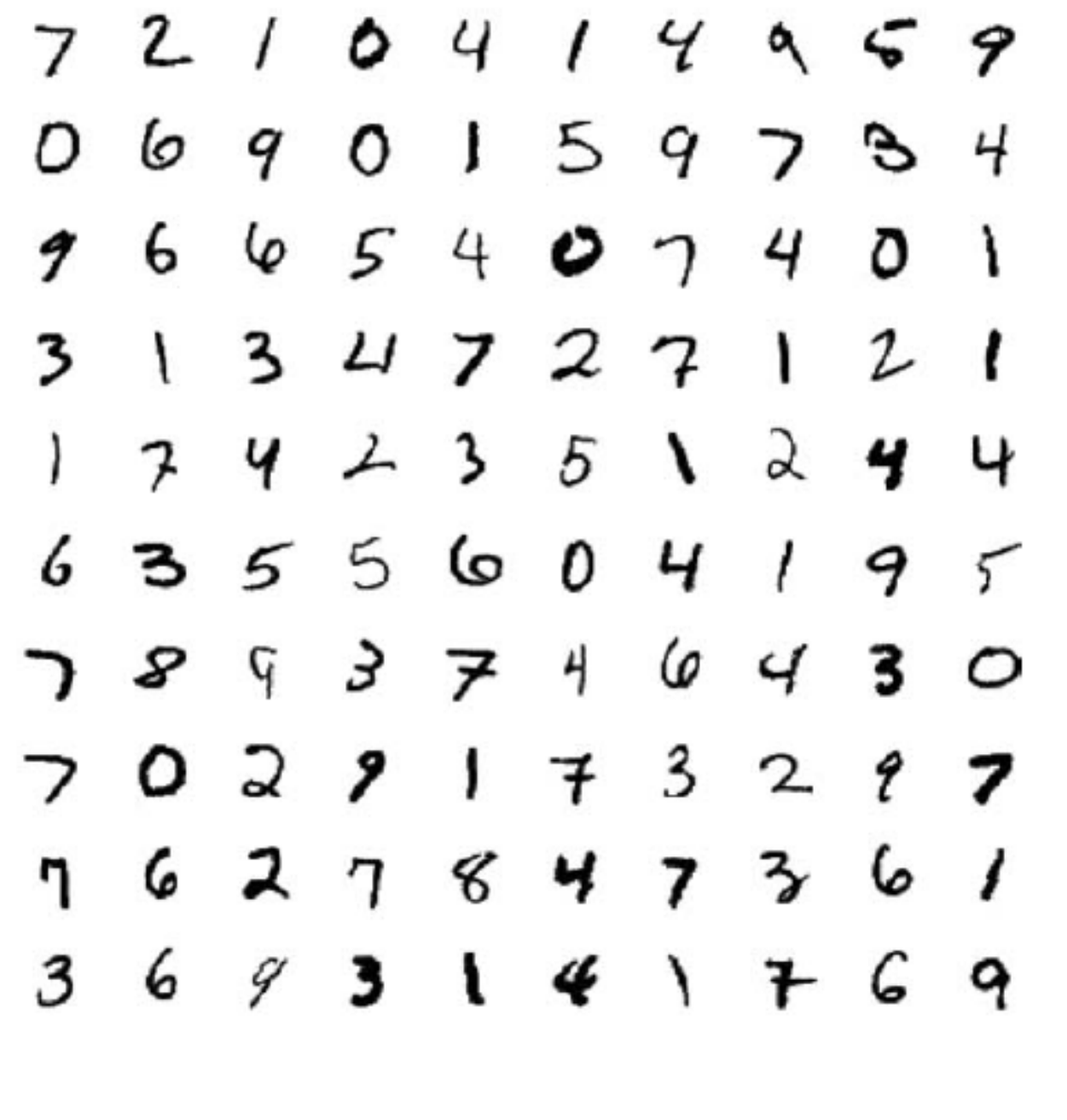}
	\caption{Sample handwritten digits from MNIST Dataset \cite{liu2003handwritten}}
	\label{fig:figure 17}
\end{figure}

The MNIST dataset is considered as one of the most used / cited dataset for handwritten digits \cite{liu2003handwritten, hangarge2010offline, cirecsan2010deep, liu2002handwritten, vamvakas2008hierarchical, babu2014handwritten}. It is the subset of  the NIST dataset and that is why it is called modified NIST or MNIST. The dataset consist of 60,000 training and 10,000 test images. Samples are normalized into 20 x 20 grayscale images with reserved aspect ratio and the normalized images are of size 28 x 28. The dataset greatly reduces the time required for pre-processing and formatting, because it is already in a normalized form.

\subsection{UCOM}
\begin{figure} [!htb]
	\centering
	\includegraphics [scale=0.6]{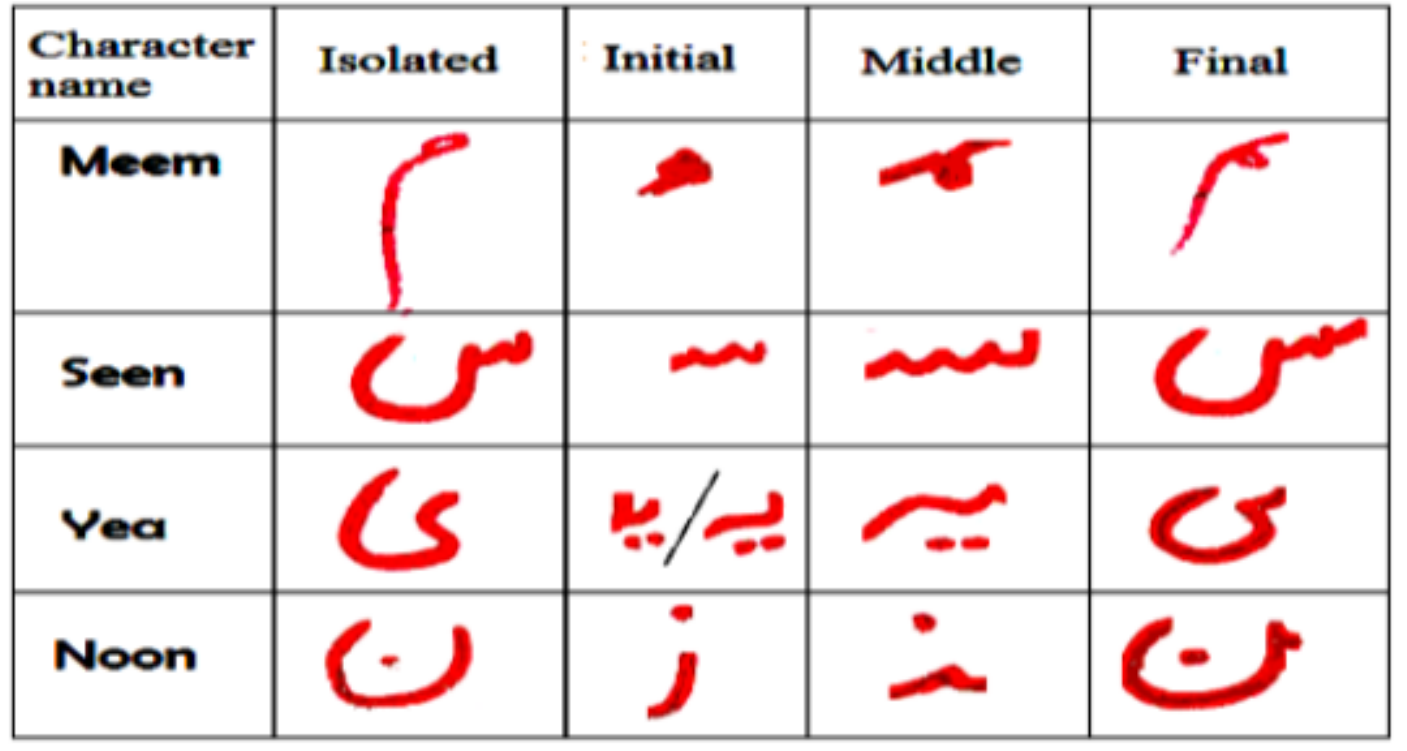}
	\caption{Example hand written characters from UCOM Dataset \cite{ahmed2017ucom}}
	\label{fig:figure 18}
\end{figure}

The UCOM is an Urdu language dataset available for research \cite{ahmed2017ucom}. The authors claim that this dataset could be used for both character recognition as well as writer identification.  The dataset consists of 53,248 characters and 62,000 words written in nasta'liq (calligraphy) style, scanned at 300 dpi. The dataset was created based on the writing of 100 different writers where each writer wrote 6 pages of A4 size. The dataset evaluation is based on 50 text line images as train dataset and 20 text line images as test dataset with reported error rate between 0.004 -0.006\%. Example characters from the dataset are presented in Figure \ref{fig:figure 18}.

\subsection{IFN/ENIT}

\begin{figure}[!htb]
	\centering
	\includegraphics [scale=0.8]{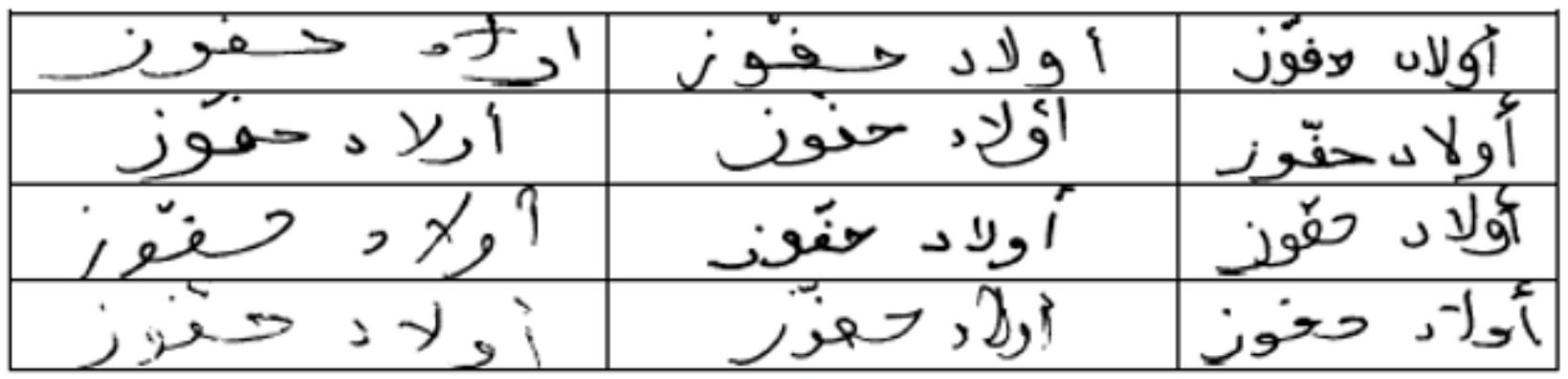}
	\caption{Sample writings from IFN/ENIT Dataset \cite{pechwitz2002ifn}}
	\label{fig:figure 19}
\end{figure}

The IFN/ENIT \cite{pechwitz2002ifn} is the most popular Arabic database of handwritten text. It was developed in 2002 by the researchers at Technical University Braunschweig, Germany for advancement of research and development of Arabic handwriting recognition systems. The dataset contains 26459 handwritten images of the names of towns and villages in Tunisia. These images consist of 212,211 characters written by 411 different writers, refer Figure \ref{fig:figure 19}. Since inception, the dataset has been widely used by the researchers for the efficient recognition of Arabic characters \cite{pechwitz2003hmm,el2007ifn,margner2007arabic,bulacu2007text}.

\subsection{CENPARMI}

The CENter for PAttern Recognition and Machine Intelligence (CENPARMI) introduced first version of Farsi dataset in 2006 \cite{khosravi2007introducing,solimanpour2006standard} . This dataset contains 18,000 samples of Farsi numerals. These numerals are divided into 11,000 training, 2,000 verification and 5,000 samples for testing purpose. 

\begin{figure} [!htb]
	\centering
	\includegraphics [scale=0.5]{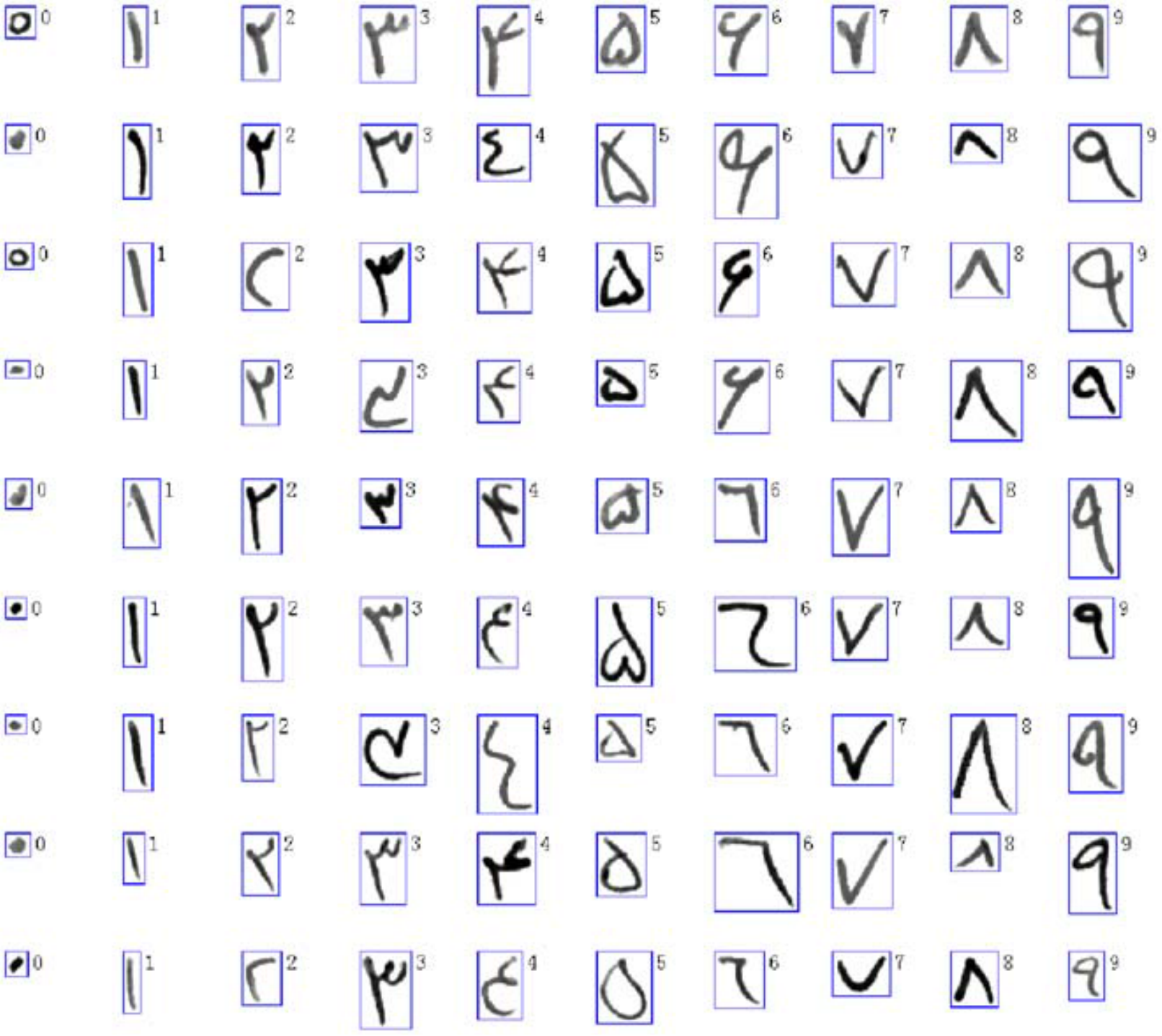}
	\caption{CENPARMI dataset example images \cite{khosravi2007introducing}}
	\label{fig:figure 20}
\end{figure}

Another similar, but larger dataset of Farsi numerals was produced by Khosravi \cite{khosravi2007introducing} in 2007. This dataset contains 102,352 digits extracted from registration forms of high school and undergraduate students. Later in 2009 \cite{haghighi2009new}, CENPARMI released another larger, extended version of Farsi dataset. This larger dataset contains 432,357 images of dates, words, isolated letters, isolated digits, numeral strings, special symbols, and documents. Refer Figure \ref{fig:figure 20} for examples images from CENPARMI Farsi language dataset.


\subsection{HCL2000}

\begin{figure}[!htb]
	\centering
	\includegraphics [scale=0.6]{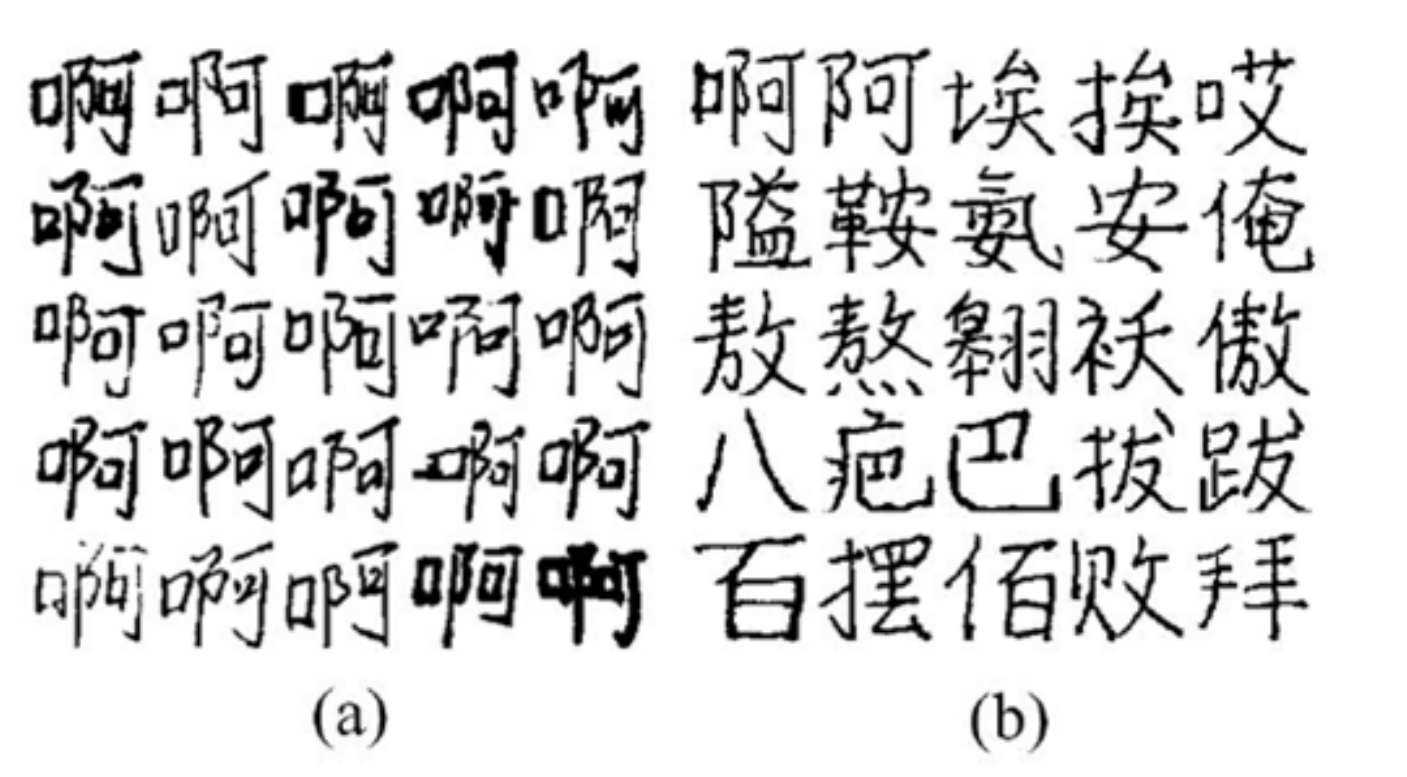}
	\caption{HCL2000 dataset sample images \cite{zhang2009hcl2000}}
	\label{fig:figure 21}
\end{figure}

The HCL2000 is an handwritten Chinese character database, refer Figure \ref{fig:figure 21} to see sample images. The dataset is publicly available for researchers. The dataset contains 3,755 frequently used Chinese characters written by 1,000 different subjects. The database is unique in a way that it contains two sub datasets, one is handwritten Chinese characters dataset, while the other is corresponding writer's information dataset. This information is provided so that research can be conducted not only based on the character recognition, but also on writer's background such as age, gender, occupation and education \cite{zhang2009hcl2000}.

\subsection{IAM}
\begin{figure}[!htb]
	\centering
	\includegraphics [scale=0.5]{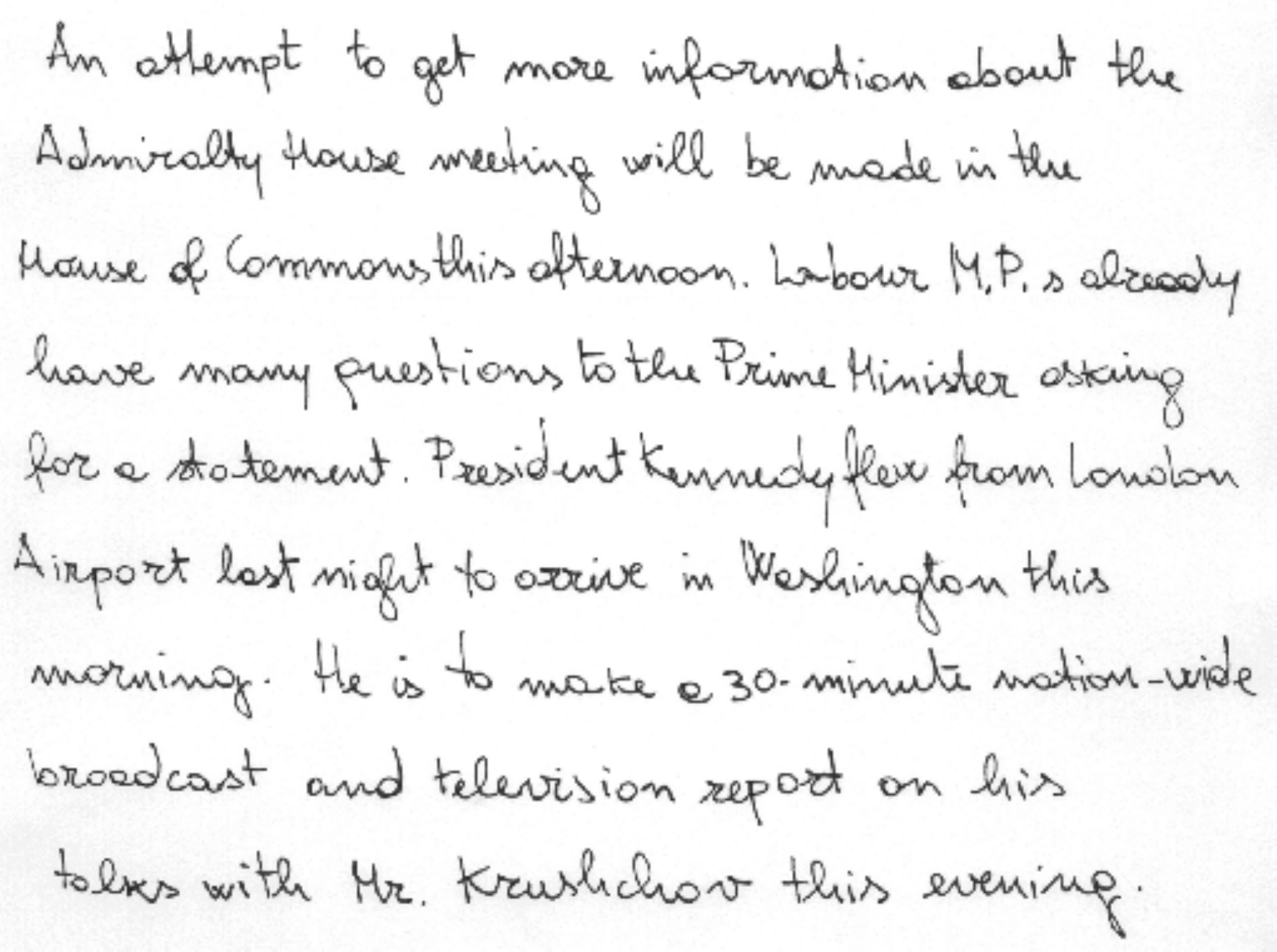}
	\caption{Sample Image IAM dataset \cite{marti2002iam}}
	\label{fig:figure 22}
\end{figure}

The IAM \cite{marti2002iam} is handwritten database of English language based on  Lancaster-Oslo/Bergen (LOB) corpus. Data were collected from 400 different writers who produced 1,066 forms of English text containing vocabulary of 82,227 words. Data consists of full English language sentences. The dataset was also used for writer identification \cite{bulacu2007text}. Researchers were able to successfully identify writer 98\% of the time during experiments on IAM dataset. Writing sample from the IAM dataset are presented in Figure  \ref{fig:figure 22}. 


\section{Languages} \label{lang}
As mentioned above, researchers working in the domain of optical character recognition have mainly investigated six different languages, which are English, Arabic, Indian, Chinese, Urdu and Persian. This is one of the future work to built OCR systems for other languages as well.

\begin{figure}[!htb]
	\centering
	\includegraphics [scale=0.78]{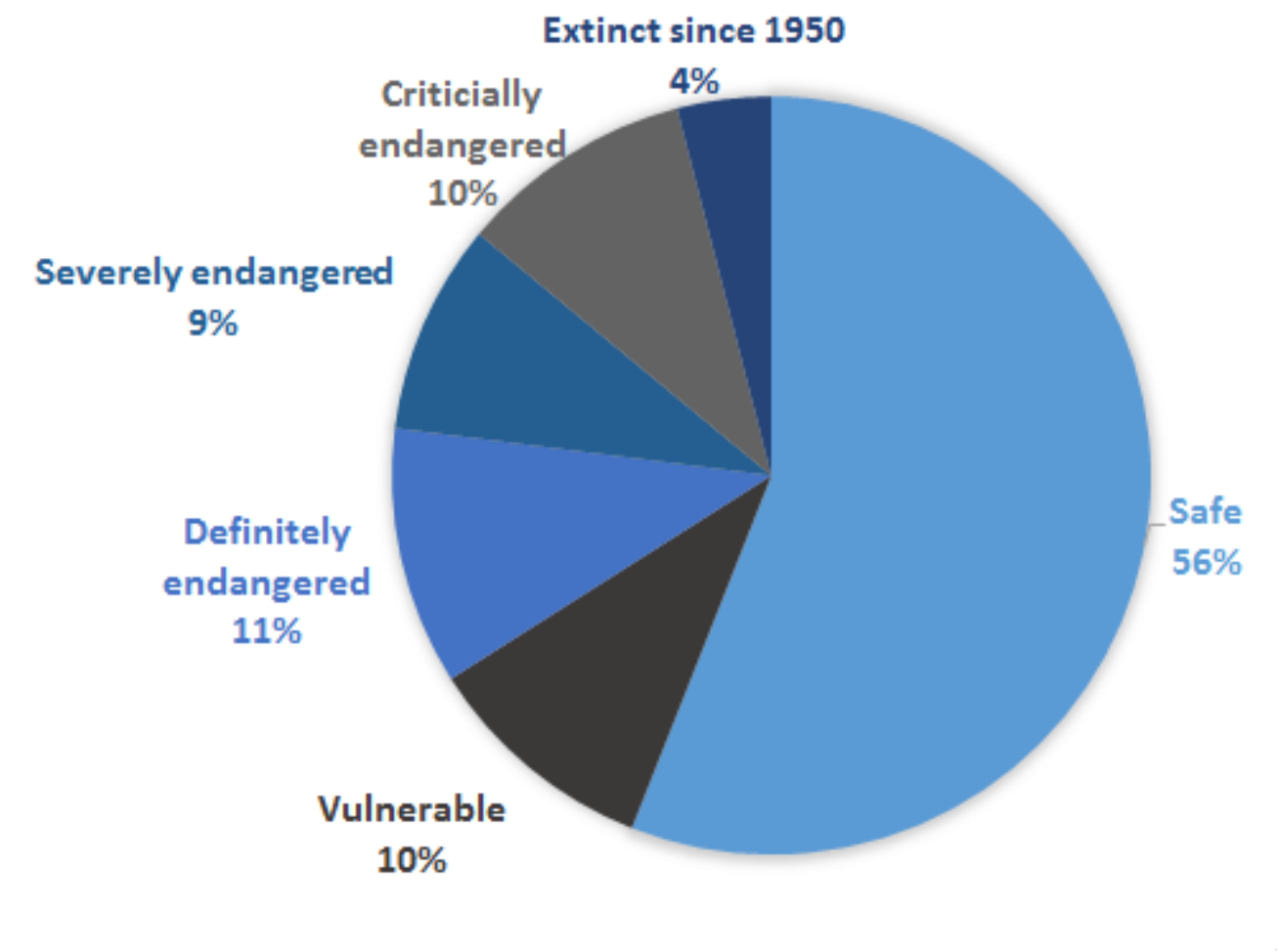}
	\caption{Data from UNESCO's report on ``world's languages in danger'' \cite{Danger}.}
	\label{fig:endang}
\end{figure}

According to the United Nations Educational, Scientific and Cultural Organization (UNESCO) report on ``world's languages in danger'' at least 43\% of languages spoken in the world are endangered \cite{Danger}. These large number of languages need attention of OCR research community as well to preserve this heritage from extinction or at least to built such system that translates documents from endangered languages to electronic form for reference.  Data from UNESCO's report on ``world's languages in danger'' is presented in Figure \ref{fig:endang}.

This section presents state-of-the art results for six language which are usually studied by researchers.

\subsection{English language}

English Language is the most widely used language in the world. It is the official language of 53 countries and articulated as a first language by around 400 million people. Bilinguals use English as an international language. Character recognition for English language has been extensively studied throughout many years. In this systematic literature review, English language has the highest number of publications i.e. 45 publications after concluding study selection process (refer Section \ref{SSP} and Section \ref{LSR}). The OCR systems for English language occupy a significant place as large number of studies have been done in the era of 2000-2018 on English language.

The English language OCR systems have been used successfully in a wide array of commercial applications. The most cited study for English language handwritten OCR is by Plamondon et al. \cite{plamondon2000online} in 2000, which have more than 2900 citations , refer Table \ref{Tab:Table 3}. The objective of the research by Plamondon et al. was to present a broad review of the state of the art in the field of automatic processing of handwriting. This paper explained the phenomenon of pen based computers and achieve the goal of automatic processing of electronic ink by mimicking and extending the pen paper metaphor. To identify the shape of the character, structural and rule based models like (SOFM) self-organized feature map, (TDNN) time delay neural network and (HMM) hidden markov model was used.


Another comprehensive overview on character recognition presented in \cite{arica2001overview} by Arica et al. has more than 500 citations. Arica et al. concluded that characters are natural entities and it is  practically impossible for character recognition to impose strict mathematical rule on the patterns of characters. Neither the structural nor the statistical models can signify a complex pattern alone. The statistical and structural information for many characters pattern can be combined by neural networks (NNs) or harmonic markov models (HMM). 

Connell et al. \cite{connell2001template} demonstrated a template-based system for online character recognition, which is capable of representing different handwriting styles of a particular character. They used decision trees for efficient classification of characters and achieve 86\% accuracy.

Every language has specific way of writing and have some diverse features that distinguished it with other language. We believe that to efficiently recognize handwritten and machine printed text of the English language, researchers have used almost all of the available feature extraction and classification techniques. These feature extraction and classification techniques include but not limited to HOG \cite{tian2016multilingual} , bidirectional LSTM \cite{toselli2016hmm}, directional features \cite{deshmukh2009analysis},  multilayer perceptron (MLP) \cite{ahlawat2017off,liu2002handwritten,sharma2013performance}, hidden markov model(HMM) \cite{zimmermann2002automatic,graves2008unconstrained,graves2009novel,pradeep2012neural}, Artificial neural network (ANN) \cite{patel2011handwritten,zhang2007novel,saha2013optical} and support vector machine (SVM) \cite{liu2005classification,vamvakas2010handwritten}. 

Recently trend is shifting away from using handcrafted features and moving towards deep neural networks. Convolutional Neural Network (CNN) architecture, a class of deep neural networks, has achieved classification results that exceeds state-of-the-art results specifically for visual stimuli / input \cite{avadesh2018optical}. LeCun \cite{DL} proposed CNN architecture based on multiple stages where each stage is further based on multiple layers. Each stage uses feature maps, which are basically arrays containing pixels. These pixels are fed as input to multiple hidden layers for feature extraction and a connected layer, which detects and classifies object \cite{KHAN201961}.


\subsection{Farsi / Persian script}
Farsi, also known as Persian Language is mainly spoken in Iran and partly in Afghanistan, Iraq, Tajikistan and Uzbekistan by approximately 120 million people. The Persian script is considered to be similar to Arabic, Urdu, Pashto and Dari languages.  Its nature is also cursive so the appearance of the letter changes with respect to positions. The script comprises of 32 characters and unlike Arabic language, the writing direction of the Farsi language is mostly but not exclusively from right to left. 

Mozaffari et. al \cite{mozaffari2004recognition} proposed a novel handwritten character recognition method for isolated alphabets and digits of Farsi and Arabic language by using fractal codes. On the basis of the similarities of the characters they categorized the 32 Farsi alphabets into 8 different classes. A multilayer perceptron (MLP) (refer Figure \ref{fig:figure 12} for overview of MLP)  was used as a classifier for this purpose. The classification rate for characters and digits were 87.26\% and 91.37\% respectively.

However, in another research \cite{soltanzadeh2004recognition}, researchers achieved recognition rate of 99.5\% by using RBF kernel based support vector machine. Broumandnia et al. \cite{broumandnia2007fast} conducted research on Farsi character recognition and claims to propose the fastest approach of recognizing Farsi character using Fast Zernike wavelet moments and artificial neural networks (ANN). This model improves on average recognition speed by 8 times. 


Liu et al. \cite{liu2009new} presented results of handwritten Bangla and Farsi numeral recognition on binary and gray scale images. The researchers applied various character recognition methods and classifiers on the three public datasets such as ISI Bangla numerals, CENPARMI Farsi numerals, and IFHCDB Farsi numerals and claimed to have achieved the highest accuracies on the three datasets i.e. 99.40\%, 99.16\%, and 99.73\%, respectively. 

In another research Boukharouba and Bennia \cite{boukharouba2017novel} proposed SVM based system for efficient recognition of handwritten digits. Two feature extraction techniques namely chain code histogram (CCH) \cite{cch} and white-black transition information were discussed. The feature extraction algorithm used in the research did not require digits to be normalized. SVM classifier along with RBF kernel method was used for classification of handwritten Farsi digits named ‘hoda’. This system maintains high performance with less computational complexity as compared to previous systems as the features used were computationally simple.

Lately, as discussed above researchers are using Convolutional Neural Network (CNN) in conjunction with other techniques for the recognition of characters. These techniques are being applied on different datasets to check the accuracy of techniques \cite{sokar2018generic, akbari2018novel, ghasemi2018persian, alizadehashraf2017persian}.

\subsection{Urdu language}

Urdu is curvasive language like Arabic, Farsi and many others \cite{naz2014optical}.  An early notable attempt to improve the methods for Urdu OCR is by Javed et al. in 2009 \cite{javed2009improving}. Their study focuses on the Nasta'liq (calligraphy) style specific pre-processing stage in order to overcome the challenges posed by the Nasta'liq style of Urdu handwriting. The steps proposed include page segmentation into lines and further line segmentation into sub-ligatures, followed by base identification and base-mark association. 94\% of the ligatures were accurately separated with proper mark association. 

Later in 2009, the first known dataset for Urdu handwriting recognition was developed at Centre for Pattern Recognition and Machine Intelligence (CENPARMI) \cite{sagheer2009new}. Sagheer et al. \cite{sagheer2009new} focused on the methods involving data collection, data extraction and pre-processing. The dataset stores dates, isolated digits, numerical strings, isolated letters, special symbols and 57 words. As an experiment, Support Vector Machine (SVM) using a Radial Base Function / kernel (RBF) was used for classification of isolated Urdu digits. The experiment resulted in a high recognition rate of 98.61\%.

To facilitate multilingual OCR, Hangarge et al. \cite{hangarge2010offline} proposed a texture-based method for handwritten script identification of three major scripts: English, Devnagari and Urdu. Data from the documents were segmented into text blocks and / or lines. In order to discriminate the scripts, the proposed algorithm extracts fine textural primitives from the input image based on stroke density and pixel density. For experiments, $k$-nearest neighbor classifier was used for classification of the handwritten scripts. The overall accuracy for tri-script and bi-script classification peaked up to 88.6\% and 97.5\% respectively.

A study by Pathan et al. \cite{pathan2012recognition} in 2012 proposed an approach based on invariant moment technique to recognize the handwritten isolated Urdu characters. A dataset comprising of 36800 isolated single and multi-component characters was created. For multi-component letters, primary and secondary components were separated, and invariant moments were calculated for each. The researchers used SVM for classification, which resulted an overall performance rate of 93.59\%. Similarly, Raza et al. \cite{raza2012unconstrained} created an offline sentence database  with automatic line segmentation. It comprises of 400 digitised forms by 200 different writers. 

Obaidullah et al. \cite{obaidullah2015numeral} proposed a handwritten numeral script identification (HNSI) framework to identify numeral text written in Bangla, Devanagari, Roman and Urdu. The framework is based on a combination of daubechies wavelet decomposition \cite{Daub} and spatial domain features. A dataset of 4000 handwritten numeral word image for these scripts was created for this purpose. In terms of average accuracy rate, multi-layer perceptron (MLP) (refer Figure \ref{fig:figure 12} for pictorial depiction of MLP) proves to be better than NBTree, PART, Random Forest, SMO and Simple Logistic classifiers.

In 2018, Asma and Kashif \cite{Asma} presented  comparative analysis of raw images and meta features from UCOM dataset. CNN (Convolutional Neural Network) and a LSTM (Long short-term Memory), which is a recurrent neural network based architecture were used on Urdu language dataset. Researchers claim that CNN provided accuracy of 97.63\% and 94.82\% on thickness graph and raw images respectively. While, the accuracy of LSTM was 98.53\% and 99.33\%. 

In another study Naseer et al. \cite{naseer2018comparative} and Tayyab et al.  \cite{tayyab2018multi} proposed an OCR model based on CNN and BDLSTM (Bi-Directional LSTM). This model was applied on dataset containing urdu news tickers and results were compared with google's vision cloud OCR. researchers found that their proposed model worked better than google's cloud vision OCR in 2 of the 4 experiments.

\subsection{Chinese language}

Our research includes 17 research publications on the OCR system of Chinese language after concluding study selection process (refer Section \ref{SSP} and Section \ref{LSR}).  One of the Earliest research on Chinese language was done in 2000 by Fu et al. \cite{fu2000user}. The researchers used self-growing probabilistic decision-based neural networks (SPDNNs) to develop a user adaptation module for character recognition and personal adaption. The resulting recognition accuracy peaked up to 90.2\% in ten adapting cycles.

Later in 2005, a comparative study of applying feature vector-based classification methods to character recognition by Cheng and Fujisawa \cite{liu2005classification} found that discriminative classifiers such as artificial neural network (ANN) and support vevtor machin (SVM) gave higher classification accuracies than statistical classifiers when sample size was large. However, in the study SVM demonstrated better accuracies than neural networks in many experiments. 

In another study Bai and Huo \cite{bai2005study} evaluated use of 8-directional features to recognize online handwritten Chinese characters. Following a series of processing steps, blurred directional features were extracted at uniformly sampled locations using a derived filter, which forms a 512-dimensional vector of raw features. This, in comparison to an earlier approach of using 4-directional features, resulted in a much better performance.  

 
In 2009, Zhang \cite{zhang2009hcl2000} presented HCL2000, a large-scale handwritten Chinese Character database. It stores 3,755 frequently used characters along with the information of its 1000 different writers. HCL2000 was evaluated using three different algorithms; Linear Discriminant Analysis (LDA), Locality Preserving Projection (LPP) and Marginal Fisher Analysis (MFA). Prior to the analysis, a Nearest Neighbor classifier assigns input image to a character group. The experimental results show MFA and LPP to be better than LDA.

Yin et al. \cite{yin2013icdar} proposed ICDAR 2013 competition which received 27 systems for 5 tasks – classification on extracted feature data, online/offline isolated character recognition and online/offline handwritten text recognition. Techniques used in the systems were inclusive of LDA, Modified quadratic discriminant function (MFQD), Compound Mahalanobis Function (CMF), convolutional neural network (CNN) and multilayer perceptron (MLP). It was explored that the methods based on neural networks proved to be better for recognizing both isolated character and handwritten text. 

During the study in 2016 on accurate recognition of multilingual scene characters, Tian et al. \cite{tian2016multilingual} proposed an extension of Histogram of Oriented Gradient (HOG), Co-occurrence HOG (Co-HOG) and Convolutional Co-HOG (ConvCo-HOG) features. The experimental results show the efficiency of the approaches used and higher recognition accuracy of multilingual scene texts. 

In 2018, researchers on Chinese script used neural networks to recognize CAPTCHA (Completely Automated Public Turing test to tell Computers and Humans Apart) recognition \cite{lin2018chinese}, Medical document recognition \cite{zhao2018multi}, License plate recognition \cite{luo2018multiple} and text recognition in historical documents \cite{yang2018recognition}. Researchers used Convolutional Neural Network(CNN) \cite{lin2018chinese,yang2018recognition}, Convolutional Recurrent Neural Network(CRNN) \cite{zhao2018multi} and Single Deep Neural Network(SDNN) \cite{luo2018multiple} during these studies.

\subsection{Arabic script}

Research on handwritten Arabic OCR systems has passed through various stages over the past two decades. Studies in the early 2000s focused mainly on the neural network methods for recognition and developed variants of databases \cite{mezghani2002line}. In 2002, Mario Pechwitz \cite{pechwitz2002ifn} developed the first IFN/ENIT-database to allow for the training and testing of Arabic OCR systems. This is one of the highly cited databases and has been cited more than 470 times. Another database was developed by Saeed Mozaffari \cite{mozaffari2006comprehensive,mozaffari2009icdar} in 2006. It stores gray-scale images of isolated offline handwritten 17,740 Arabic / Farsi numerals and 52,380 characters. Another notable dataset containing Arabic handwritten text images was introducted by Mezghani et al. \cite{mezghani2012database}. The dataset has open vocabulary written by multiple writers (AHTID / MW). It can be used for word and sentence recognition, and writer identification \cite{khayyat2014learning}.

A survey by Lorigo and Govindaraju \cite{lorigo2006offline} provides a comprehensive review of the Arabic handwriting recognition methodologies and databases used until 2006. This includes research studies carried out on IFN/ENIT database. These studies mostly involved artificial neural networks (ANNs), Hidden Markov Models (HMM), holistic and segmentation-based recognition approaches. The limitations pointed out by the review included restrictive lexicons and restrictions on the text appearance. 


In 2009, Alex et al. \cite{graves2009offline} introduced a globally trained offline handwriting recognizer based on multi-directional recurrent neural networks and connectionist temporal classification. It takes raw pixel data as input. The system had an overall accuracy of 91.4\% which also won the international Arabic recognition competition. 

Another notable attempt for Arabic OCR was made by Lutf et al. \cite{lutf2014arabic} in 2014, which primarily focused on the specialty of Arabic writing system. The researcher proposed a novel method with minimum computation cost for Arabic font recognition based on diacritics. Flood-fill based and clustering based algorithms were  developed for diacritics segmentation. Further, diacritic validation is done to avoid misclassification with isolated letters. Compared to other approaches, this method is the fastest with an average recognition rate of 98.73\% for 10 most popular Arabic fonts.

 An Arabic handwriting synthesis system devised by Elarian et al. \cite{elarian2015arabic} in 2015 synthesizes words from segmented characters. It uses two concatenation models: Extended-Glyphs connection and the Synthetic-Extensions connection. The impact of the results from this system shows significant improvement in the recognition performance of an HMM based Arabic text recognizer. 

Hicham and Akram \cite{akram2016using} discussed an analytical approach to develop a recognition system based on HMM Toolkit (HTK). This approach requires no priori segmentation. Features of local densities and statistics are extracted using vertical sliding windows technique, where each line image is transformed into a series of extracted feature vectors. HTK is used in the training phase and Viterbi algorithm is used in the recognition phase. The system gave an accuracy of 80.26\% for words with “Arabic-numbers” database and 78.95\% with IFN / ENIT database.

In study conducted in 2016 by Elleuch et al. \cite{elleuch2016new}, convolutional neural network (CNN) based on support vector machine (SVM) is explored for recognizing offline handwritten Arabic. The model automatically extracts features from raw input and performs classification.  

In 2018, researchers applied technique of DCNN (deep CNN) for recognizing the offline and handwritten Arabic characters \cite{boufenar2018investigation}. An accuracy of 98.86\% was achieved when strategy of DCNN using transfer learning was applied on two datasets. In another similar study \cite{jebril2018recognition} an OCR technique based on HOG (Histograms of Oriented Gradient) \cite{ICME_Khan} for feature extraction and SVM for character classification was used on handwritten dataset. The dataset contained names of Jordanian cities, towns and villages yielded an accuracy of 99\%. However,  when the researchers used multichannel neural network for segmentation and CNN for recognition on machine printed characters, the experiments on 18pt font showed an overall accuracy of 94.38\%.

\subsection{Indian script}
Indian script is collection of scripts used in the sub-continent namely Devanagari \cite{avadesh2018optical}, Bangla \cite{rabby2018bornonet}, Hindi \cite{dutta2017towards}, Gurmukhi \cite{singh2011feature}, Kannada \cite{sagar2008ocr} etc. One of the earliest research on Devanagari (Hindi) script was proposed in 2000 by Lehal and Bhatt \cite{lehal2000recognition}. The research was conducted on Devanagari script and English numerals. The researchers used data that was already in isolated form in order to avoid the segmentation phase. The research is based on statistical and structural algorithms \cite{kimura1991handwritten}. The results of Devanagari scripts were better than English numerals. Devanagari had recognition rate of 89\% with 4.5 confusion rate, while English numerals had recognition rate of 78\% with confusion rate of 18\%. 

Patil et. al \cite{patil2002neural} was the first researcher to use neural network approach for the identification of Indian documents. The researchers propose a system capable of reading English, Hindi and Kannada scripts. Modular neural network was used for script identification while a two stage feature extraction system was developed, first to dilate the document image and second to find average pixel distribution in the resulting images. 

Sharma et al. \cite{sharma2006recognition} proposed a scheme based on quadratic classifier for the recognition of Devanagari script. The researchers used 64 directional features based on chain code histogram \cite{cch} for feature recognition. The proposed scheme resulted in 98.86\% and 80.36\% accuracy in recognizing Devanagari characters and numeral respectively. Fivefold cross validation was used for the computation of results.

Two research studies \cite{hanmandlu2007fuzzy,hanmandlu2007input} presented in 2007 were based on use of fuzzy modeling for character recognition of Indian script. The researchers claim that the use of reinforcement learning on a small database of 3500 Hindi numerals helped achieve recognition rate of 95\%.

Another research carried out on Hindi numerals \cite{bhattacharya2009handwritten} used relatively large dataset of 22,556 isolated numeral samples of Devanagari and 23,392 samples of Bangla scripts. The researchers used three Multi-layer perceptron classifiers to classify the characters. In case of a rejection, a 4th perceptron was used based on the output of previous three perceptrons in a final attempt to recognize the input numeral.  The proposed scheme provided 99.27\% recognition accuracy vs the fuzzy modeling technique, which provided the accuracy of 95\%.

Desai \cite{desai2010gujarati} used neural networks for the numeral recognition of Gujrati script. The researcher used a multi-layers feed forward neural network for the classification of digits. However, the recognition rate was low at 82\%. 

Kumar et al. \cite{garg2010s,garg2010new} proposed a method for line segmentation of handwritten Hindi text. An accuracy of 91.5\% for line segmentation and 98.1\% for word segmentation was achieved. Perwej et. al \cite{perwej2012machine} used back propagation based neural network for the recognition of handwritten characters. The results showed that the highest recognition rate of 98.5\% was achieved. Obaidullah et al. \cite{obaidullah2015numeral} proposed Handwritten Numeral Script Identification or HNSI framework based on four indic scripts namely, Bangla, Devanagari, Roman and Urdu. The researchers used  different classifiers namely NBTree, PART, Random Forest, SMO, Simple Logistic and MLP and evaluated the performance against the true positive rate. Performance of MLP was found to be better than the rest. MLP was then used for bi and tri-script identification. Bi-script combination of Bangla and Urdu gave the highest accuracy rate of 90.9\% on MLP, while the highest accuracy rate of 74\% was achieved in tri-script combination of Bangla, roman and Urdu.

In a multi dataset experiment\cite{rabby2018bornonet}, researchers applied a lightweight model based on 13 layers of CNN with 2-sub layers on four datasets of Bangla language.  An accuracy of 98\%, 96.81\%, 95.71\%, and 96.40\% was achieved when model was applied on CMATERdb, ISI, BanglaLekha-Isolated dataset and mixed datasets respectively. CNN based model was also applied on ancient documents written in Devanagari or Sanskrit script in another study. Results, when compared with Google's vision OCR gave an accuracy of 93.32\% vs 92.90\%.

\section{Research trends} \label{future}

Lately, the research in the domain of optical character recognition has moved towards deep learning approach \cite{naz2017urdu,al2018deep} with little to no emphasis on hand crafted features. In this section we have analyze research trend / techniques mainly used in the publications of last three years (2015-2018). Our analysis is summarized in Table \ref{Tab:Table 4}. 

Table \ref{Tab:Table 4} includes script under investigation, techniques or classification technique employed for OCR, year of publication and respective reference number. This table gives holistic view of how researchers working on some of the widely used languages are trying to solve the problem of optical character recognition. We can see that neural network, specially CNN is being used extensively for the recognition of optical characters. However, traditional techniques like SVM, HMM, SIFT etc. are also being used in conjunction with CNN. 

\begin{landscape}
	\begin{longtable}{| p{1.5cm} | p{16cm}| p{1.5cm}| p{1cm}|}

		\toprule
		\label{Tab:Table 4}
		\textbf{Script}  &\textbf{Technique Employed}& \textbf{Year} &\textbf{ Ref} \\ \hline	
		
		\endhead
		\multicolumn{3}{@{}l}{\ldots \textit{continued on next page}}
		
		\endfoot
		\endlastfoot

Chinese English Indian & Two new feature descriptors Co-HoG and ConvCo-HoG based on Histogram of Oriented Gradient(HoG) based on CNN.  & 2016 & \cite{tian2016multilingual}\\ \hline
Chinese & Convolutional Neural Network(CNN), Long Short-Term Memory (LSTM), Hidden Markov Model(HMM) & 2016 & \cite{suryani2016benefits}\\ \hline
Chinese & Neural Network language model,  Convolutional Neural Networks  & 2017 & \cite{wu2017improving}\\ \hline
Chinese English Indian & BOW based representation for characters, DSN-SVM,HOG-SVM, FV-SVM  & 2017 & \cite{shi2017fisher}\\ \hline
Chinese English & Convolutional neural network (CNN) & 2017 & \cite{feng2017robust}\\ \hline
Chinese & Convolutional Neural Network (CNN) & 2018 & \cite{lin2018chinese}\\ \hline
Chinese & Convolutional-Recurrent Neural Network (CRNN) & 2018 & \cite{zhao2018multi}\\ \hline
Chinese & Single Deep Neural Network (SDNN) & 2018 & \cite{luo2018multiple}\\ \hline
Chinese & Recognition of Chinese Text in Historical Documents with Page-Level Annotations. CNN, CNN followed by LSTM & 2018 & \cite{yang2018recognition}\\ \hline

English Urdu Indian & Features: Discrete wavelet transforms (DWT), Daubechies wavelet transforms, textural and entropy features. Classifiers: NBTree, PART, LIBLinear, Random Forest, SMO, MLP & 2015 & \cite{obaidullah2015numeral}\\ \hline

English Indian & SVM and Shortest Path Algorithm &2017 &\cite{chaudhuri2017approach}\\ \hline

English & Heat Kernel Signature (HKS), HKS with SIFT and triangular mesh structure &2015 &\cite{zhang2015handwritten}\\ \hline

English & Long Short-Term Memory (LSTM) architecture & 2015 & \cite{gregor2015draw}\\ \hline

English & word graphs (WG) based Line-level keyword spotting (KWS). & 2016 & \cite{toselli2016hmm}\\ \hline

English & Recurrent Neural Network (RNN)  model with two multi-layer RNNs trained with Long Short Term Memory (LSTM) and connectionist temporal classification (CTC), CNN & 2017 & \cite{su2017accurate}\\ \hline
English & Box Approach, Mean, Standard Deviation, Centre of Gravity, Neural Network & 2017 & \cite{ahlawat2017off}\\ \hline

Arabic & Kashida features, width feature and Hidden Markov Model (HMM) & 2015 & \cite{elarian2015arabic}\\ \hline
Arabic & Elliptic grapheme codebook features, $X^2$ distance metric & 2015 & \cite{abdi2015model}\\ \hline
Arabic & Features of local densities and features statistics, Hidden Markov Model (HMM) & 2016 & \cite{akram2016using}\\ \hline
Arabic & Support Vector Machine(SVM), Convolutional Neural Network(CNN) & 2016 & \cite{elleuch2016new}\\ \hline
Arabic & Recurrent connectionist language modeling 	 &2017 &\cite{yousfi2017contribution}\\ \hline

Arabic & DCNN (Deep Convolutional neural network) & 2018 & \cite{boufenar2018investigation}\\ \hline

Arabic & Edge operators (Canny, Sobel, Prewitt, Roberts, and Laplacian of Gaussian), Transforms (Fourier (FFT), discrete cosine (DCT), Hough, and Radon), Texture features (Gray-level range and standard deviation, entropy of the gray-level distribution, and the properties of the gray-level co-occurrence matrix (GLCM)), Moments (Hu’s seven moments and Zernike moments) and Ensemble of Support Vector Machine(SVM) & 2018 & \cite{elanwar2018making}\\ \hline

Arabic & Histograms of Oriented Gradient (HOG) , SVM & 2018 & \cite{jebril2018recognition}\\ \hline
Arabic & Multi-Channel Neural Network (MCNN) & 2018 & \cite{radwan2018neural}\\ \hline
Arabic & Fast Automatic Hashing Text Alignment (FAHTA) &2018 & \cite{doush2018novel}\\ \hline
Arabic & Deep Siamese Convolutional Neural Network and SVM & 2018 & \cite{sokar2018generic}\\ \hline

Urdu & Hierarchical combination of Convolutional Neural Networks (CNN),  Multi-dimensional Long Short-Term Memory Neural Networks (MDLSTM)  & 2017 & \cite{naz2017urdu}\\ \hline
Urdu & BDLSTM (Bi-Directional Long Short-Term Memory),  Recurrent Neural Network (RNN) & 2018 & \cite{tayyab2018multi}\\ \hline
Urdu & Histogram of Oriented Gradient (HOG), Support Vector Machine (SVM), $k$ Nearest Neighbors ($k$NN), Random Forest (RF) and Multi-Layer Perceptron (MLP) & 2018 & \cite{chandio2018character}\\ \hline
Urdu & Convolutional Neural Network (CNN) and Long Short-Term Memory (LSTM) & 2018 & \cite{naseer2018comparative}\\ \hline


Indian &  Multi-column Multi-scale Convolutional Neural Network (MMCNN)   &2017 &\cite{sarkhel2017multi}\\ \hline
Indian & Convolutional Neural Network (CNN) & 2018 & \cite{rabby2018bornonet}\\ \hline
Indian & Convolutional Neural Network (CNN) & 2018 & \cite{avadesh2018optical}\\ \hline
Indian & Histogram of oriented gradient (HOG) and Support Vector Machine (SVM) & 2018 & \cite{choudhury2018handwritten}\\ \hline
Indian & Convolutional Recurrent Neural Network (CRNN) with Spatial Transformer Network (STN) layer & 2018 & \cite{dutta2017towards}\\ \hline
Indian & Zoning, Discrete Cosine Transformations (DCT), gradient features, $k$ Nearest Neighbors ($k$NN), Support Vector Machine (SVM), Decision Tree and Random Forest & 2018 & \cite{kumar2018improved}\\ \hline

Persian & Chain Code Histogram (CCH), transition information in the vertical and horizontal directions,  Support Vector Machine(SVM) & 2017 & \cite{boukharouba2017novel}\\ \hline


Persian & Zoning, chain code, outer profile, crossing count, $k$ Nearest Neighbors ($k$NN), Artificial Neural Networks and Support Vector Machine (SVM) & 2018 & \cite{akbari2018novel}\\ \hline
Persian & Convolutional Neural Network (CNN) & 2018 & \cite{sarvaramini2018persian}\\ \hline
Persian & Convolutional Neural Network (CNN) & 2018 & \cite{ghasemi2018persian}\\ \hline

\caption{Summary of frequently used feature extraction and classification techniques: Data corresponding to last three years (2015-2018). Studies corresponding to ``Indian'' script do include research on scripts belonging to Devanagari, Bangla, Hindi, Gurmukhi, Kannada etc. }

\end{longtable}
\end{landscape}

\section{Conclusion and future work} \label{conclusion}

\subsection{Conclusion}

\begin{enumerate}
\item Optical character recognition has been around for last eight (8) decades. However, initially products that recognize optical characters were mostly developed by large technology companies. Development of machine learning and deep learning has enabled individual researchers to develop algorithms and techniques, which can recognize handwritten manuscripts with greater accuracy. 

\item In this literature review, we systematically extracted and analyzed research publications on six widely spoken languages. We explored that some techniques perform better on one script than on another e.g. multilayer perceptron classifier gave better accuracy on Devanagri and Bangla numerals \cite{bhattacharya2009handwritten, mozaffari2004recognition} but gave average results for other languages \cite{ahlawat2017off, liu2002handwritten, sharma2013performance}. The difference may have been due to the fact that how specific technique models different style of characters and quality of the dataset. 

\item Most of the published research studies propose solution for one language or even subset of a language. Publicly available datasets also include stimuli that are aligned well with each other and fail to incorporate examples that corresponds well with real life scenarios i.e. writing styles, distorted strokes, variable character thickness and illumination \cite{long2018}. 

\item  It was also observed that researchers are increasingly using Convolutional Neural Networks(CNN) for the recognition of handwritten and machine printed characters. This is due to the fact that CNN based architectures are well suited for recognition tasks where input is image. CNN were initially used for object recognition tasks in images e.g. the ImageNet Large Scale Visual Recognition Challenge (ILSVRC) \cite{ILSVRC15}. AlexNet \cite{Krizhevsky1}, GoogLeNet \cite{7298594} and ResNet \cite{ResNet} are some of the CNN based architectures widely used for visual recognition tasks.

\end{enumerate}

\subsection{Future work} \label{fworks}

\begin{enumerate}

\item As mentioned in Section \ref{lang}, research in OCR domain is usually done on some of the most widely spoken languages. This is partially due to non-availability of datasets on other languages. One of the future research direction is to conduct research on languages other than widely spoken languages i.e. regional languages and endangered languages. This can help preserve cultural heritage of vulnerable communities and will also create positive impact on strengthening global synergy.

\item Another research problem that needs attention of research community is to built systems that can recognize on screen characters and text in different conditions in daily life scenarios e.g. text in captions or news tickers, text on sign boards, text on billboards etc. This is the domain of ``recognition / classification / text in the wild''. This is complex problem to solve as system for such scenario needs to deal with background clutters, variable illumination condition, variable camera angles, distorted characters and variable writing styles \cite{long2018}.

\item To build robust system for ``text in the wild'', researchers needs to come up with challenging datasets that is comprehensive enough to incorporate all possible variations in characters. One such effort is \cite{Yuan2019}. In another attempt, research community has launched ``ICDAR 2019: Robustreading challenge on multi-lingual scene text detection and recognition'' \cite{2019arXiv190700945N}. Aim of this challenge is invite research studies that proposes robust system for multi-lingual text recognition in daily life or ``in the wild'' scenario. Recently report for this challenge has been published and winner methods for different tasks in the challenge are all based on different deep learning architectures e.g. CNN, RNN or LSTM.

\item Published research studies have proposed various systems for OCR but one aspect that needs to improve is commercialization of research. Commercialization of research will help building low cost real-life systems for OCR that can turn lots of invaluable information into searchable / digital data \cite{comm2008}.

\end{enumerate}






\pagestyle{plain} 

\appendix
 

\end{document}